\documentclass[lettersize,journal]{IEEEtran}
\usepackage{amsmath,amsfonts}
\usepackage{algorithmic}
\usepackage{algorithm}
\usepackage{array}
\usepackage{float}
\usepackage[caption=false,font=normalsize,labelfont=sf,textfont=sf]{subfig}
\usepackage{textcomp}
\usepackage{stfloats}
\usepackage{url}
\usepackage{verbatim}
\usepackage{graphicx}
\usepackage{cite}
\usepackage{xcolor}
\usepackage{gensymb}
\usepackage{booktabs}
\usepackage{multirow}
\usepackage{threeparttable}
\usepackage{pifont}

\begin{document}

\title{AllTact Fin Ray: A Compliant Robot Gripper with Omni-Directional Tactile Sensing}

\author{Siwei Liang, Yixuan Guan, Jing Xu, Hongyu Qian, Xiangjun Zhang, Dan Wu, Wenbo Ding, Rui Chen
\thanks{Siwei Liang, Yixuan Guan, Jing Xu contributed equally to this work. (Corresponding author: Rui Chen, E-mail: chenruithu@tsinghua.edu.cn). }
\thanks{Siwei Liang, Yixuan Guan, Jing Xu, Hongyu Qian, Xiangjun Zhang, Dan Wu, and Rui Chen are with the Department of Mechanical Engineering, Tsinghua University, Beijing, China.}
\thanks{Wenbo Ding is with Tsinghua-Berkeley Shenzhen Institute, Institute of Data and Information, Shenzhen International Graduate School, Tsinghua University, Shenzhen, China.}
}

\maketitle

\begin{abstract}
Tactile sensing plays a crucial role in robot grasping and manipulation by providing essential contact information between the robot and the environment. In this paper, we present AllTact Fin Ray, a novel compliant gripper design with omni-directional and local tactile sensing capabilities.
The finger body is unibody-casted using transparent elastic silicone, and a camera positioned at the base of the finger captures the deformation of the whole body and the contact face.
Due to the global deformation of the adaptive structure, existing vision-based tactile sensing approaches that assume constant illumination are no longer applicable. To address this, we propose a novel sensing method
where the global deformation is first reconstructed from the image using edge features and spatial constraints. Then, detailed contact geometry is computed from the brightness difference against a dynamically retrieved reference image.
Extensive experiments validate the effectiveness of our proposed gripper design and sensing method in contact detection, {force estimation}, object grasping, and precise manipulation.
\end{abstract}

\def\abstractname{Note to Practitioners}
\begin{abstract}
Fin Ray fingers have been widely adopted for robot grasping and manipulation due to their adaptive nature.
However, existing Fin Ray finger designs lack tactile sensing capabilities, or can only perceive contacts at certain regions.
The motivation of this work is to enable the Fin Ray gripper to not only detect contacts in all directions but also measure the detailed contact geometry of the grasped object. 
The omni-directional contact sensing ability is crucial for manipulation in cluttered and dynamic scenarios. And the contact geometry measurement is critical for in-hand pose estimation and precise manipulation. 
The proposed finger design is only composed of a single transparent elastic silicone cast, a camera and white LEDs, making the manufacturing process simple and time-efficient. The design files and sensing algorithms will be open-sourced to foster further development.
We validate the effectiveness of this sensor design and sensing method across various applications. It can locate contact positions with an error less than 1mm, {accurately estimate the force magnitude on the contact surface}, and detect contact forces from different directions with a threshold below 2 N. In future research, we will explore advanced manipulation applications including deformable object and bi-manual manipulation and improve the finger's durability.

\end{abstract}


\section{Introduction}
\label{sec:new_intro}

\begin{figure}[t]
\centering  
    \centering
    \subfloat[]{
        \includegraphics[width=0.95\linewidth]{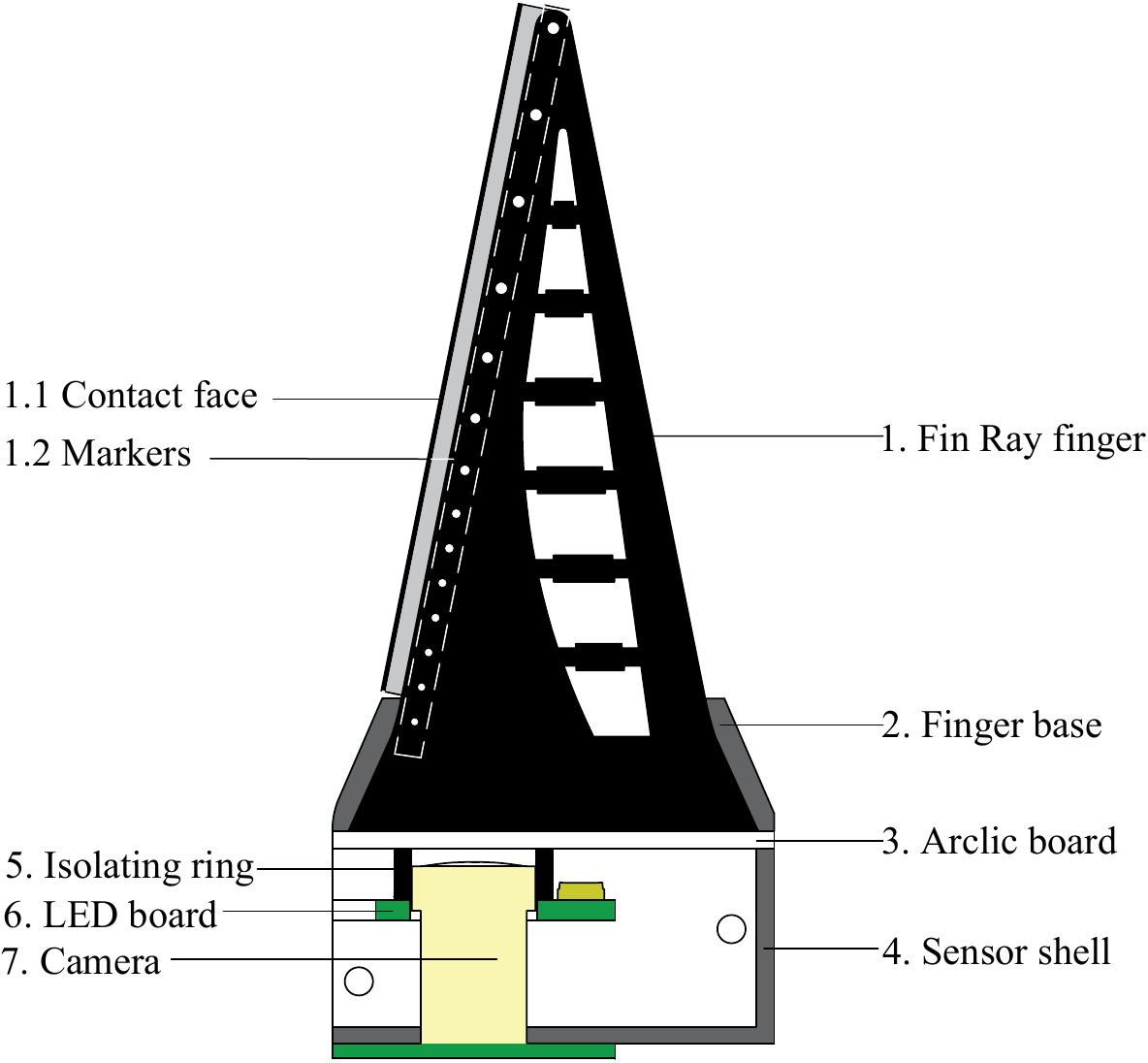}
        \label{fig:cutaway}}
    \hfill
    \subfloat[]{
        \includegraphics[width=0.95\linewidth]{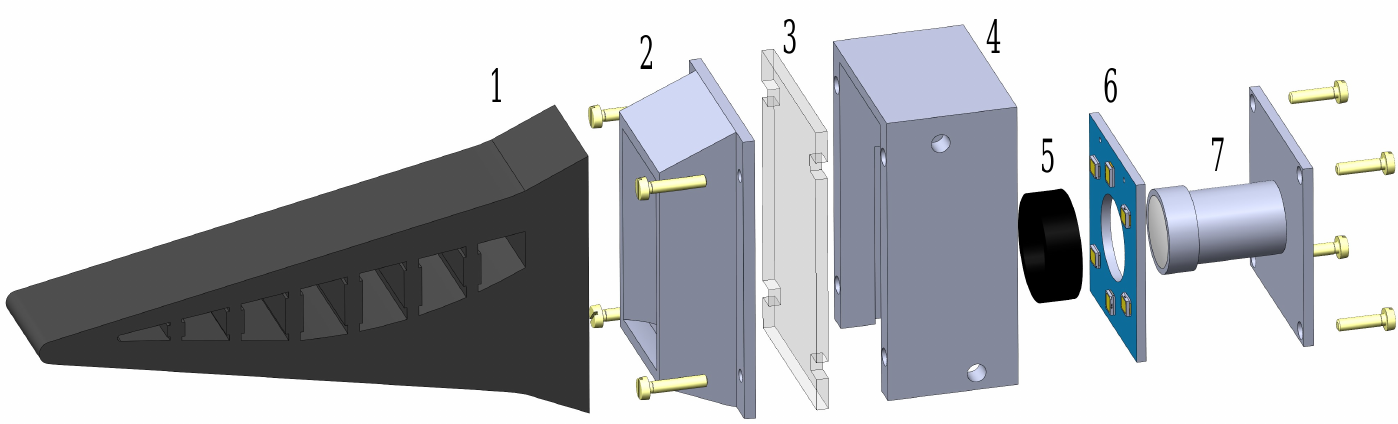}
        \label{fig:expoded}}
\caption{AllTact Fin Ray finger structure view. (a) Simplified cutaway view where the Fin Ray finger body is not cut. (b) Complete exploded view.}
\label{cutaway and exploded}
\end{figure}
\IEEEPARstart{T}{actile} sensing can provide contact information which is necessary for robot grasping and 
manipulation~\cite{Gelsight, dong2021tactile,10086535, 10636851, 10505922,chen2024tactilesim2real}.
However, due to perception errors or disturbances during execution, the robot may interact with environment using areas lacking tactile sensing feedback unexpectedly, resulting in failed execution or damage to the end effector.
{Therefore, to enable safe and precise robotic manipulation, a gripper is expected to: (1) exhibit global compliance; (2) reconstruct the overall shape of the compliantly deformed finger (global deformation); (3)  capture detailed surface contact geometry (local deformation); (4) detect contacts from all directions. 
}{Visuotactile sensing offers high spatial resolution, low manufacturing costs, and simplified wiring compared to other tactile sensing techniques, making it well-suited for integration into grippers~\cite{li2024vision}. Numerous efforts have been undertaken to design grippers with visuotactile sensing capabilities. We summarize some of these contributions in Table~\ref{tab:comparison}. However, most existing designs incorporate a rigid support for the gel, which restricts compliance. Additionally, some designs are compliant in only one direction, limiting their ability to detect contact to the designated contact face and rendering them unable to sense unexpected contacts from other directions. Although the JamTac system \cite{tata, li2023jamtac} meets all the requirements, its lack of precise control limits its suitability for robotic manipulation.
}

{
In this article, we propose AllTact Fin Ray gripper, a compliant, simple-structure gripper embedded visuotactile sensor. 
To the best of our knowledge, the gripper is the first two-fingered gripper to simultaneously satisfy all the requirements.}
The structure of AllTact Fin Ray is shown in Fig.~\ref{cutaway and exploded}.
The finger body is compliant and Fin Ray-shaped, such that it can actively bend toward the object and thereby enhance grasp stability.
In order to capture the local contact geometry at the gripper's contact face, 
we cast the finger body using transparent elastic silicone and put the rigid camera at the base of the finger.
Moreover, we use a semi-transparent layer to transform the local contact into image intensity changes such that the illumination configuration can be significantly simplified as white LEDs.
We further propose a 3D contact sensing algorithm that is able to reconstruct the body's global deformation and local contact geometry from the captured image. 
The global deformation is first computed using the camera pinhole model and geometry constraints. To address the influence of global deformation on local geometry reconstruction, we use learning-based marker detection and tracking to dynamically retrieve the corresponding reference image from a recorded video. We also design a corresponding calibration method to acquire the mapping function, where only a ball of known radius is used.

\begin{table*}[t]
\centering
\begin{threeparttable}
\caption{{Comparison of Visuotactile Fingers and Grippers}}
\color{black}{
\begin{tabular}{c|c|c|c|c|c|c}
\toprule

\multirow{2}{*}{\textbf{Name}} & \multirow{2}{*}{\textbf{Structural Compliance}} & \multirow{2}{*}{\textbf{Illumination}} & \multicolumn{3}{c|}{\textbf{Reconstruction}} & \multirow{2}{*}{\textbf{Contact Detection}} \\
\cline{4-6}
& & & \textbf{Methods} & \textbf{Local} & \textbf{Global} & \\
\hline

GelTip \cite{gomes2020geltip} & Rigid Base, Soft Pad & White & Perspective Model & Qualitative & Rigid & Omni\\
Rigid Sensors \cite{omitact,sun2022soft,tippur2023gelsight360,azulay2023allsight,do2023densetact,lambeta2024digitizingtouchartificialmultimodal} & Rigid Base, Soft Pad & RGB & PSA & \checkmark & Rigid & Omni \\
\cite{de2021external} and \cite{xu2021external} & 3D Printed FRS & — & Side-Located Camera & \ding{53} & \checkmark & Contact Face \\
GelSight Fin Ray \cite{Gelsight_finray,Gelsight_baby_finray} & Assembled FRS & RGB & PSA & \checkmark & \ding{53} & Contact Face \\
GelSight Endoflex \cite{liu_tendon} & UA Linkage, Soft Pad & RGB & ResNet & \checkmark & \ding{53} & Contact Face \\
Ma. \cite{Ma_tendon} & UA Linkage, Soft Pad & RGB & PSA & \checkmark & \checkmark & Contact Face \\
GelSight Svelte \cite{Gelsight_svelte} & Gel Skin and Backbone & RG & PSA \& CNN & \checkmark & Qualitative & Contact Face \\
JamTac \cite{tata, li2023jamtac} & Granular Package & RGB & PSA \& CNN & \checkmark & \checkmark & Omni \\
Seethrufinger \cite{see_thru_finger,see_thru_finger2,see_thru_finger3} & Soft Open-Frame & — & ResNet & \ding{53} & \checkmark & Omni \\
Work by \cite{guo2024reconstructing} & Soft Open-Frame & — & Geometric Optimization & Localization & \checkmark & Contact Face\\
TacPalm \cite{zhang2024tacpalm} & Pneumatic, Tactile Palm & White & TacTip \cite{ward2018tactip} \& CNN & \checkmark & \ding{53} & Palm only \\
GelSight FlexiRay \cite{wang2024gelsight_flexiray} & Assembled FRS & RGB & PP-LiteSeg \& ResNet & \checkmark & Qualitative & Contact Face \\
DexiTac \cite{lu2024dexitac} & Pneumatic Arm, Rigid Tip & RGB & DigiTac \cite{lepora2022digitac} & \checkmark & \ding{53} & Contact Face \\

\midrule
\textbf{AllTact Fin Ray (Ours)} & \textbf{Unibody-Cast Elastic FRS} & \textbf{White} & \textbf{Diff-Mapping \& CNN} & \textbf{\checkmark} & \textbf{\checkmark} & \textbf{Omni} \\

\bottomrule

\end{tabular}
}
\begin{tablenotes}
\small
\item PSA refers to Photometric Stereo Algorithm; FRS refers to Fin Ray Structure; UA refers to Underactuated.
\end{tablenotes}
\end{threeparttable}
\label{tab:comparison}
\end{table*}

We first verify the effectiveness of the finger design in contact {localization}, pose estimation, force detection, and local geometry reconstruction through experiments.
We further assemble two fingers into a robot gripper and mount it onto a robotic arm. The gripper retains the adaptiveness inherent to Fin Ray grippers during grasping. And by leveraging the tactile sensing signals, we can accurately estimate the in-hand object pose and adjust it accordingly.

The contributions of the study are summarized as follows.
\begin{enumerate}[] 
\item We propose a novel compliant, simple-structure Fin Ray gripper embedded visuotactile sensor that can not only detect contacts in all directions but also reconstruct detailed local contact geometry.

\item {We propose an algorithm that reconstructs the global deformation using the camera pinhole model and geometry constraints.}

\item {We propose an algorithm that reconstructs local contact geometry details. To eliminate the influence of global deformation on local geometry reconstruction, we build a mapping function from the normalized brightness difference against a dynamically retrieved reference image, and develop a calibration method to acquire the mapping.}

\item {We design an algorithm that can detect the contact in all the directions and estimate the force magnitude for contacts on the contact face.}

\item We conduct experiments and ablation studies to verify the effectiveness of our design in various applications, including contact {localization}, contact detection, object grasping and manipulation.
\end{enumerate}
We will open-source all the gripper design files and sensing algorithms upon acceptance.

The rest of the article is organized as follows. In Section~\ref{sec:related}, we introduce related works on soft grippers with tactile sensing and omni-direction visuotactile sensor. 
In Section~\ref{sec:design}, the design and fabrication procedure of the gripper are described.
In Section~\ref{sec:sensing_principle}, the tactile sensing pipeline and calibration method are presented.
In Section~\ref{sec:experiments}, experimental details and results are presented. 
Finally in Section~\ref{sec:conclusion}, we conclude our work and discuss limitations.

\section{Related works}
\label{sec:related}

Tactile sensors can be categorized into piezoresistive, capacitive, piezoelectric, fiber Bragg grating, and visuotactile (vision-based tactile) types based on their operating principles. We refer to \cite{VTS_in_soft_gripper} for a comprehensive survey. 
Visuotactile sensing offers several advantages over other tactile sensing techniques, including high spatial resolution, low manufacturing costs, and simplified wiring requirements. These characteristics make it well-suited for integration into soft grippers. Accordingly, this section reviews relevant works on visuotactile sensors and fingers.

{
\subsection{Global Deformation Reconstruction}

For soft grippers, the ability to reconstruct the global deformation of the finger during manipulation is critical for precise manipulation. External photography is the most direct way to capture global deformation. \cite{de2021external} and \cite{xu2021external} use a side-located camera to reconstruct the global deformation of the gripper, and compute contact forces by combining finite element method or kinematic model with deep learning. However, external cameras cannot capture local geometry and are not applicable in low-visibility environments.

\cite{see_thru_finger,see_thru_finger2,see_thru_finger3} introduce a Fin-Ray-like soft network-structure finger with a camera mounted below the finger to capture the global deformation. The open-frame structure allows the camera to capture both global finger deformation and external scenes. However, the open frame lacks the capability to grasp small size objects and reconstruct fine local geometries, which restricts the applications of the gripper.
\cite{guo2024reconstructing} presents a similar framework with a fully enclosed contact face, enabling both localization of local contact positions and reconstruction of global deformations, but lacks the ability to reconstruct fine local object geometry.

Gelsight FlexiRay \cite{wang2024gelsight_flexiray} develops a proprioception model based on a ResNet50 backbone, which not only qualitatively infers the global deformation, but also computes local geometry and contact force. Similarly, Gelsight Svelte \cite{Gelsight_svelte} qualitatively characterizes global deformation by tracking LED markers to estimate applied torques. However, both of the studies fail to  quantitatively reconstruct the global deformation of the fingers, while our finger overcome the limitation by computing from a single image using edge features of the image and a spatial constraint, allowing real-time visualization and monitoring of global deformation.
 }

{
\subsection{Local Geometry Reconstruction}

Local geometry reconstruction is important for in-hand object recognition and precise manipulation. The most common geometry reconstruction method for tactile sensing is Photometric Stereo Algorithm (PSA), which uses three-colored lights to illuminate the contact face, computes depth gradients from the RGB image, and reconstructs the 3D geometry from these gradients. 
The algorithm is first adopted in GelSight\cite{Gelsight,early_Gelsight}, and has been widely applied in soft visuotactile grippers.

For underactuated fingers, Liu et al. \cite{liu_tendon} install GelSight sensors in each knuckle to enable contact geometry reconstruction. Ma \cite{Ma_tendon} further improve the design by using a single camera to capture all three knuckles through placed mirrors. However, these approaches complicate the finger structure with LEDs and cameras while splitting the contact area, resulting in incomplete reconstruction.

For granular jamming grippers\cite{jamming_gripper}, Li et al. \cite{tata, li2023jamtac} soak particles in a liquid with same refractive index to eliminate the visual boundary of particles in captured images, and use PSA to reconstruct the contact geometry. 
However, the particles filled in grippers reduce the reconstruction resolution, and the suitability of jamming grippers in robot manipulation is limited.

For Fin Ray finger, Liu et al. \cite{Gelsight_finray} remove the middle part of the finger and mount a camera on the opposite of the contact face.
\cite{Gelsight_baby_finray} further improves the design by using flexible mirrors and silicon adhesive-based fluorescent paint, which improves the finger compliance. However, the semi-rigid backing and mirror limit the compliance, and they can only detect contact at the front face.

While the aforementioned grippers can successfully reconstruct local contact geometries using PSA, the algorithm’s requirement for uniform RGB illumination presents a significant integration challenge for compliant fingers, resulting in complicated structural designs. In contrast, DTact \cite{Dtact} reconstructs contact geometry by measuring changes in brightness of a semi-transparent layer during contact, simplifying illumination requirements. However, its assumption of constant illumination can not be satisfied for soft grippers. Our work overcomes this limitation by developing a mapping function that uses normalized brightness differences against dynamically retrieved reference images, effectively eliminating the influence of global deformation on local geometry reconstruction. We further develop a calibration method to establish this mapping relationship.
}

{
\subsection{Omni-Directional Contact Detection}
Omni-directional contact detection is necessary for safe manipulation\cite{omitact,sun2022soft,tippur2023gelsight360,azulay2023allsight,do2023densetact,gomes2020geltip,lambeta2024digitizingtouchartificialmultimodal}. \cite{omitact} proposes a multi-directional high-resolution tactile sensor, where multiple cameras are used to capture the contacts in different directions. However, the utilization of multiple cameras leads to increased cost and fabrication complexity. In \cite{sun2022soft}, a thumb-shaped tactile sensor is proposed, where a monocular camera is used and PSA combined with structured light is adopted for contact detection. In \cite{tippur2023gelsight360}, the authors propose a generalized cross-illumination PSA scheme and design a more compact omni-directional visuotactile sensor. In \cite{azulay2023allsight}, the authors add markers on the surface, such that the tangential forces and torsion can be detected. In \cite{do2023densetact}, a dense randomized pattern is used to eliminate the effects of aliasing in deflection tracking. Most recently, GelSight and Meta AI develop Digit 360~\cite{lambeta2024digitizingtouchartificialmultimodal}, a multi-modal omni-directional tactile sensor with a hemispherical shape. It can measure contact geometry, normal and shear forces, perceive vibrations, and sense temperature and odor. It utilizes a dynamic illumination system composed of 8 controllable RGB LEDs. Edge AI is included in the sensor to process the data and reduce the bandwidth.

The above works incorporate rigid frames or shells, which restrict their compliance and limit their ability to effectively grasp objects of varying shapes and sizes. In contrast, our approach employs a Fin Ray-shaped, unibody-casted elastomer as the finger body.  Additionally, it can detect contacts from all directions based on marker displacement.

}

\section{Gripper Design}
\label{sec:design}
\subsection{Finger Design}
\label{section2.1}

Fig. \ref{cutaway and exploded} shows a simplified cutaway view where the Fin Ray body is not cut for viewing the markers and a complete exploded view of our finger and sensor. 

The finger consists of a Fin Ray finger body, a rigid base, a camera and white LEDs. 
The whole finger body is cast in transparent elastic silicone such that it can deform and fit to the object during grasping and manipulation. The unibody casting design also significantly simplifies the sensor fabrication process. 

In order to capture the contact face using the camera at the bottom of the finger, we change the original design of the Fin Ray finger and thicken the contact-side.
An extra layer of semi-transparent silicone is cast above the contact face for local geometry reconstruction. The surfaces of the finger body are further painted black to block the environmental illumination. The reconstruction principle will be described in Sec.~\ref{sec:sensing_principle}.
Unlike PSA that requires multi-color illumination, our sensor utilizes simple white LEDs for illumination. Additionally, we incorporate an isolating ring between the camera and LEDs to prevent direct light from the LEDs from entering the camera, ensuring accurate imaging.

The side faces feature dotted markers along the edges, which are tracked during sensing to obtain the reference image corresponding to real-time image and accomplish reconstruction.
While similar markers are used in \cite{Gelsight_finray}, ours are strategically placed on the side faces to prevent interference with the contact image {and provide  force sensing ability}. Due to the perspective projection of the imaging process, the radius and spacing of the markers increase proportionally with their distance from the camera, ensuring relatively uniform marker sizes in the captured image.

\subsection{Finger Fabrication}
\label{section2.2}
\begin{figure}[!t]
\centering
\includegraphics[width=\linewidth]{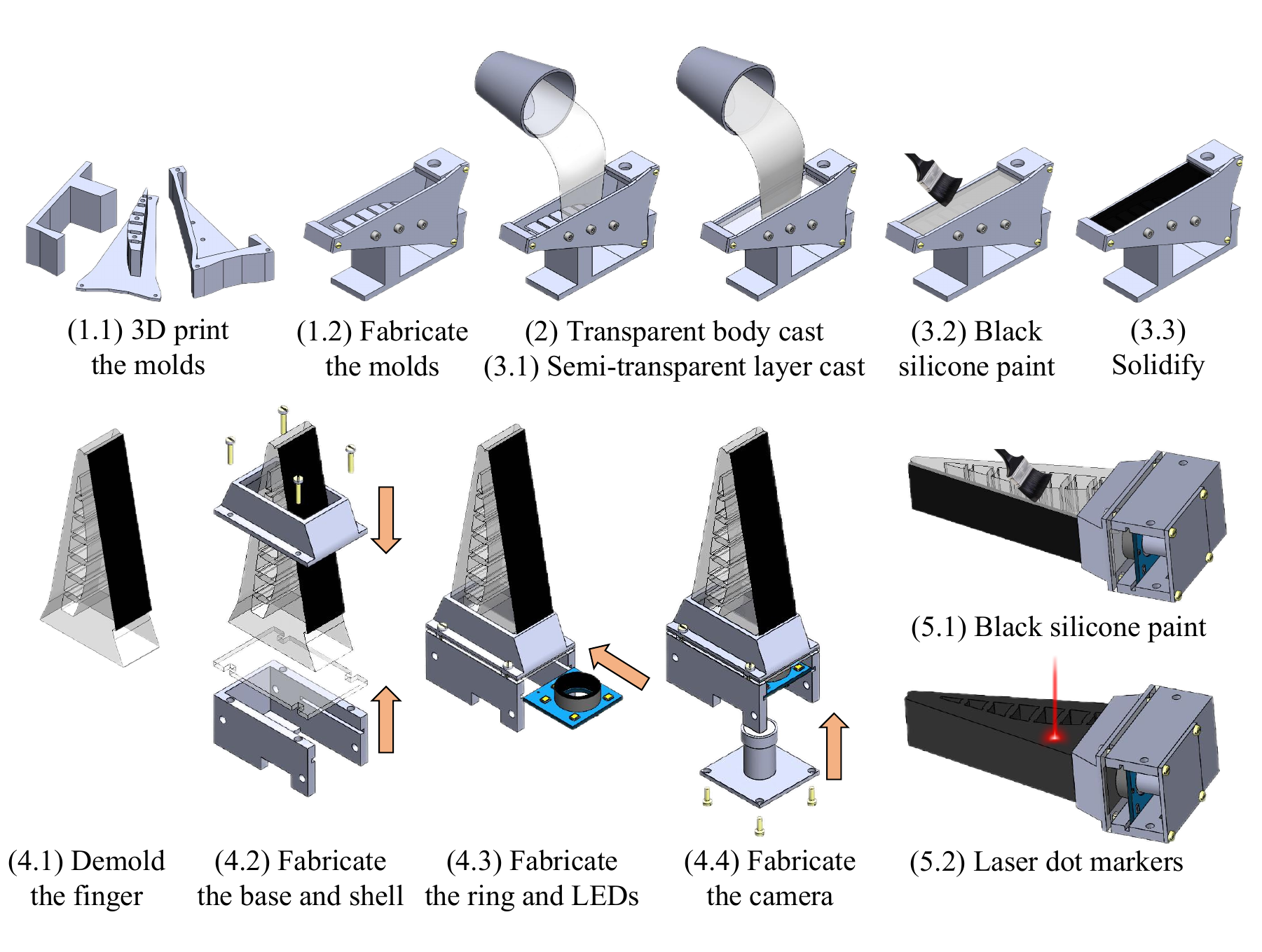}
\caption{{Diagram of the Fabrication Process for AllTact Fin Ray Finger}}
\label{fig:fabrication}
\end{figure}

The fabrication of AllTact Fin Ray finger involves 5 processes {as shown in Fig. \ref{fig:fabrication}} and follows:

\begin{enumerate}[] 
\item 
\textit {Manufacture the Components: }
{
The components shown in Fig. \ref{cutaway and exploded} include the finger base and sensor shell, which are 3D-printed using white resin (Greatsimple Technology$^\circledR$ UTR8360), and the isolating ring, which is 3D-printed using black resin (Somos$^\circledR$ Taurus). The LED board and acrylic board are custom-designed and fabricated. Additionally, the casting molds for the Fin Ray finger are 3D-printed with white resin.
}

\item 
\textit {Unibody Cast the Finger: }
We use transparent elastic silicone (ELASTOSIL$^\circledR$ RT 601, 9:1 ratio, {Shore A hardness 40}) for the finger body,
which is harder and less locally deformed than the semi-transparent layer when contacting.
The silicone is mixed thoroughly before being poured into the mold.

\item
\textit{Cast the Contact Face: }
After the finger body has solidified, we cast the contact face onto it. 
We first cast a {1.5 mm thick} semi-transparent silicone layer on the transparent body (POSILICONE$^\circledR$ , ratio A:B=1:1, {Shore A hardness 5}), {which is soft and sensitive to contact.} After the layer has solidified, the surface is painted with the same silicone {dyed black} (Sic Pig$^\circledR$ black). Finally, we demold the finger after all the materials have solidified. Since we directly cast the contact layer onto the silicone finger body, a robust bond forms during solidification, which is essential for maintaining integrity when the finger deforms during contact. 

\item 
\textit {Assemble: }
First, the finger body is adhered to the base using glue. The finger base is then secured to the sensor shell with screws, with the acrylic board fixed between them. Next, the LED board, glued with the isolating ring, is inserted into slots on the sensor shell. Finally, the camera is screwed to the sensor shell. The camera has a resolution of 1920 $\times$ 1080 pixels, a frame rate of 30 FPS, and a field of view (FOV) of 85$\degree$.

\item 
\textit {Surface Process: }
After the assembly, the whole finger is coated with black paint to eliminate the influence of the environmental light. 
In order to retrieve the reference image for local geometry reconstruction, we use a laser engraving machine to create dot markers on both side faces of the finger.

\end{enumerate}

The unibody casting design significantly simplifies the fabrication process and shortens the time cost. The whole manufacture time without the solidification is less than an hour.

\begin{figure}[!t]
    \centering
    \subfloat[]{
        \includegraphics[width=0.40\linewidth]{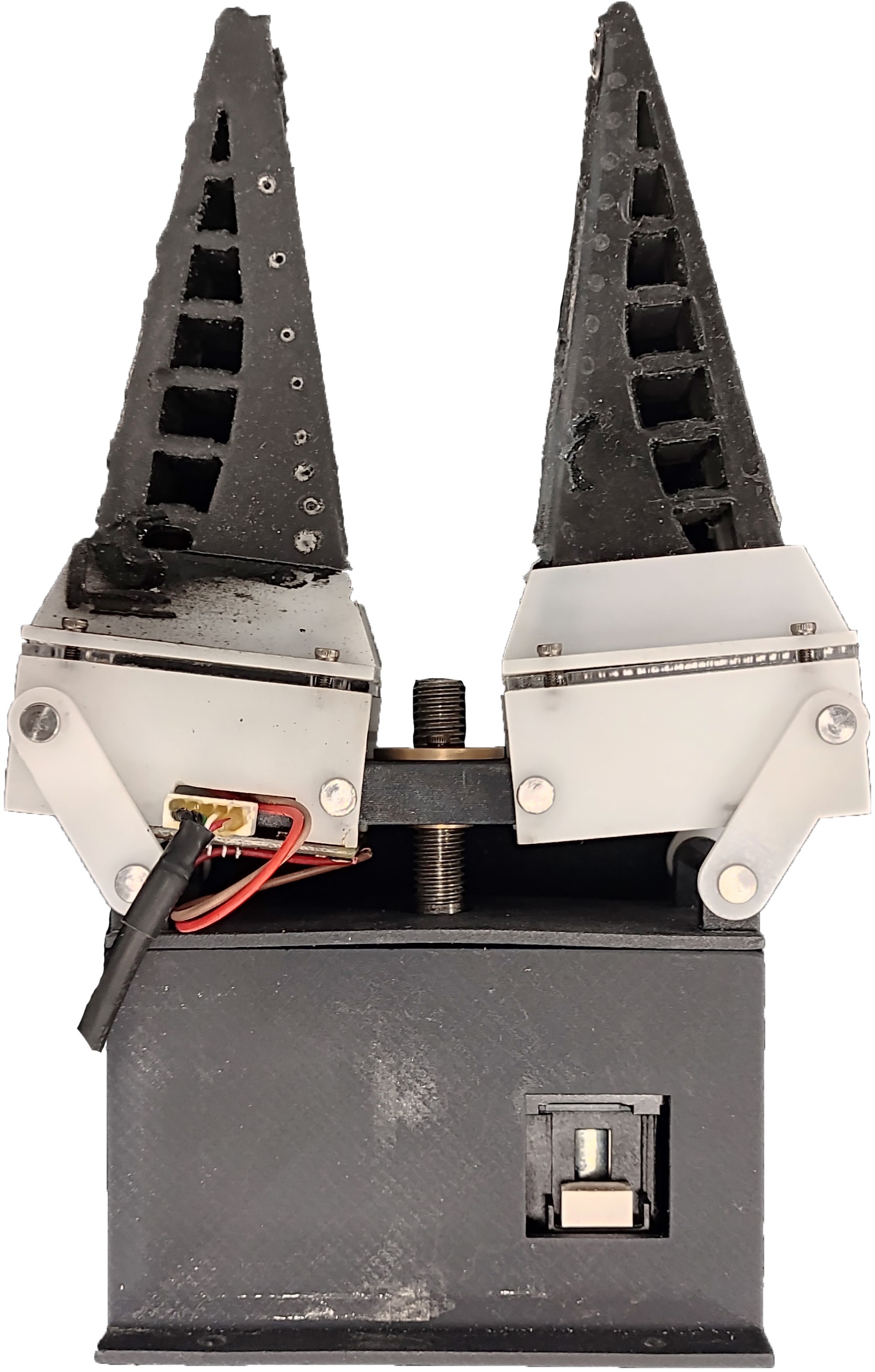}
        \label{fig:gripper in model}}
    \subfloat[]{
        \includegraphics[width=0.58\linewidth]{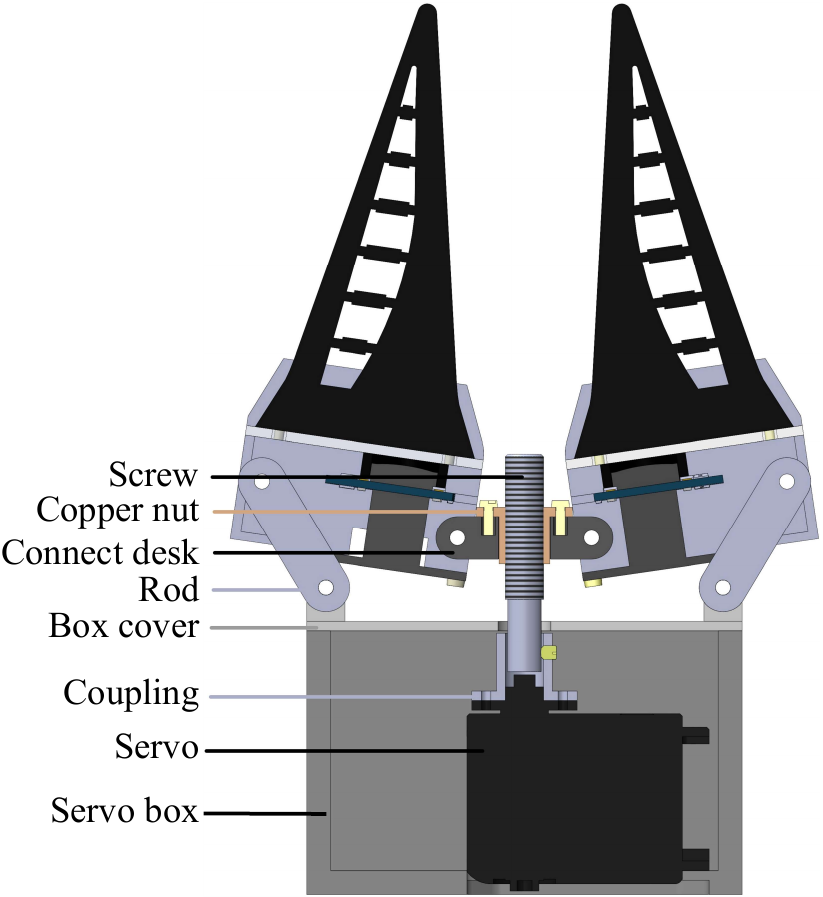}
        \label{fig:physical gripper}}
\caption{Full view of AllTact Fin Ray gripper. (a) View of physical gripper. (b) Cutaway view of gripper in CAD model.}
\label{fig:gripper_structure}
\end{figure}
For grasping and manipulation, we further assemble two fingers into a gripper.
Fig. \ref{fig:gripper_structure} shows a full view of the gripper.
The gripper is driven by a rod-and-screw mechanism, offering self-locking capabilities and enabling precise control of finger movement. 
The actuator we employ is an electromotor Dynamixel MX-28 AT. The gripper includes Fin Ray fingers, rods, a connection desk, a copper nut, an M8 screw, a motor box, and a box cover. Pins are used to connect the fingers to the connection desk, the fingers to the rods, and the rods to the holes in the box cover.

\section{Sensing Principles}
\label{sec:sensing_principle}
\begin{figure}[]
\centering  
\includegraphics[width=\linewidth]{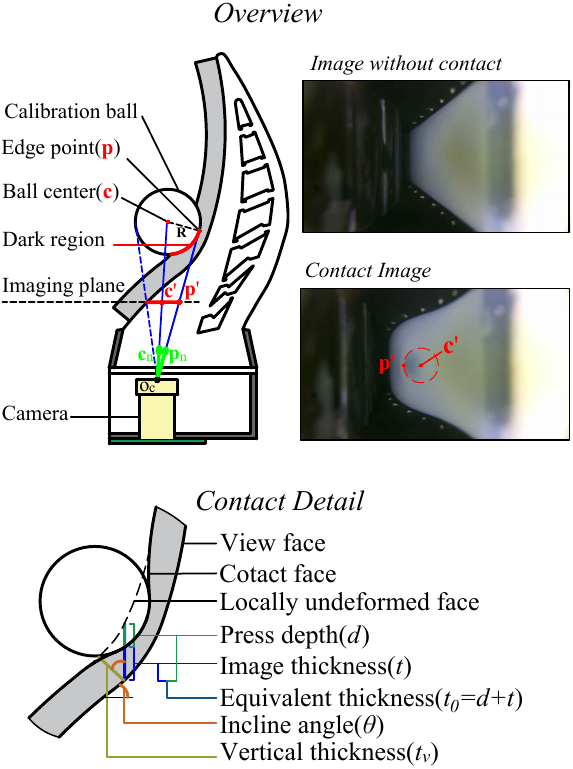}
\caption{Diagram of deformation and calibraion of our finger during contact, where a ball with known radius is pressed onto the contact face for sensor calibration. The camera pinhole model is employed to compute the 3D coordinate of ${\bf{c}}$, which is further used to calculate the press depth.}
\label{fig:principle}
\end{figure}

{
The sensing principles of the sensor consist of three modules:
 global deformation reconstruction (Sec.~\ref{sec:global_deformation}), local geometry reconstruction (Sec.~\ref{sec:local_reconstruction}), and contact force detection (Sec.~\ref{sec:contact_detection}). Fig.~\ref{fig:principle} illustrates the diagram and definitions of each term.

}

\subsection{Global Deformation Reconstruction}
\label{sec:global_deformation}
We propose an algorithm that infers the global deformation of the \textit{view face} in the 3D space from a single image using image edge features and spatial constraint. 

\begin{figure}[!t]
\centering
\includegraphics[width=\linewidth]{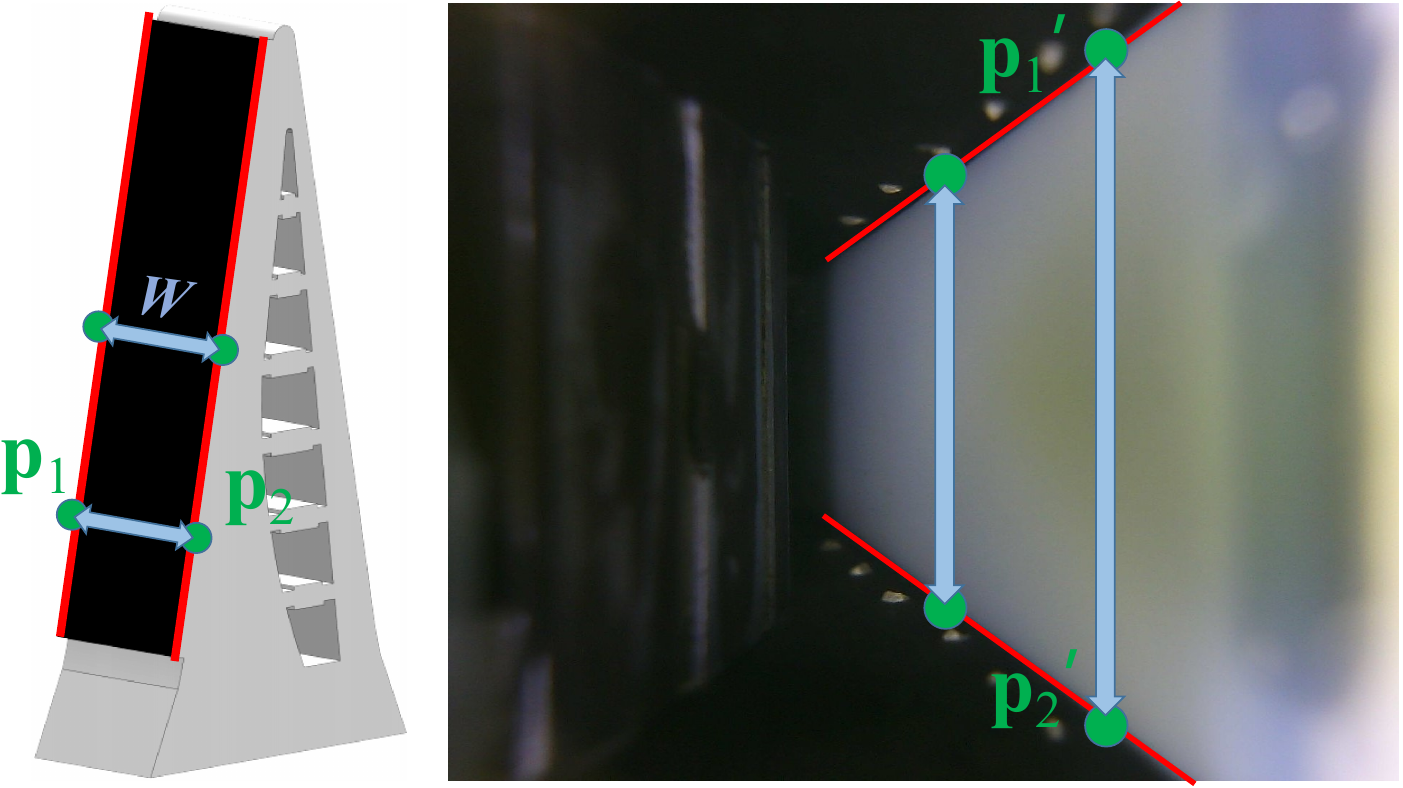}
\caption{Diagram of global deformation reconstruction. We use $W$ and the image coordinates ${{\bf{p}}'_1}$ and ${{\bf{p}}'_2}$ {on the two edges} to compute the depth and 3D coordinates of points. {Note that ${{\bf{p}}_1}$ and ${{\bf{p}}_2}$ are not markers doted on side faces but drawn on the two edges of the contact face.}}
\label{fig:global}
\end{figure}

According to the pinhole camera model, the coordinate transformation relationship without distortion is formed as:

\begin{equation}
\label{pinhole imaging}
z{\bf{p}}' ={\bf{K}}{\bf{p}}
\end{equation}
where ${\bf{K}} \in \mathbb{R}^{3 \times 3}$ is the intrinsic matrix of the camera whose parameters can be obtained through calibration, ${\bf{p}}'$ and ${\bf{p}}$ are the image pixel coordinate and the 3D coordinate in the camera frame, respectively:
\begin{equation}
{\bf{p}}' = (u,v,1)^\top\ \ \ \ \ \ \ {\bf{p}} = (x,y,z)^\top 
\end{equation}
As $z$ is unknown, each pixel in the image corresponds to a line in the 3D space. To compute ${\bf{p}}$, we {first localize the edges of contact region in the image through binarization and edge detection in OpenCV and }impose an additional constraint: 
for one point ${\bf{p}}_{1}$ on one edge of the finger, its distance to the corresponding point on the opposite edge ${\bf{p}}_{2}$ equals to the finger's width $W$, as shown in Fig. \ref{fig:global} and the equation:
\begin{equation}
    \left\|{\bf{p}}_{1} -{\bf{p}}_{2}\right\|_2 = W
    \label{eq:distance_constraint}
\end{equation}
When mounting the finger body, we align its width direction with the camera coordinate system's $y$-axis and image's $v$-axis. Therefore, the corresponding points are defined as {points sharing identical image pixel coordinates $u$}:
\begin{equation}
    u_1 = u_2
    \label{eq:u_constraint}
\end{equation}
Furthermore, the lines connecting these corresponding points $\overrightarrow{{\bf{p}}_{1}{\bf{p}}_{2}}$ are horizontal: 
\begin{equation}
    z_{1} = z_{2}
    \label{eq:z_constraint}
\end{equation}
By combining (\ref{eq:distance_constraint}), (\ref{eq:u_constraint}) and (\ref{eq:z_constraint}), we can compute the $z$ coordinates of ${\bf{p}}_{1}$ and ${\bf{p}}_{2}$. Then we can derive the 3D coordinates from (\ref{pinhole imaging}) for all the points on edges.
For the points between two edges, a linear interpolation is applied. Hence, we {can localize} all the points on the \textit{view face} and reconstruct the global deformation. {The algorithmic framework for global deformation reconstruction is shown in Algorithm~\ref{alg:global_deformation}.}

\begin{algorithm}[t]
\renewcommand{\algorithmicrequire}{\textbf{Input:}}
\renewcommand{\algorithmicensure}{\textbf{Output:}}
\caption{\textcolor{black}{Global deformation reconstruction}}\label{alg:global_deformation}
\begin{algorithmic}[1]
\REQUIRE \textcolor{black}{Grayscale image $I$ from camera}
\ENSURE \textcolor{black}{Locally undeformed face $\mathcal{P}_{LUF}$}
\textcolor{black}{
    \STATE Localize edges $\mathcal{E}_1$, $\mathcal{E}_2$ in $I$
    \FOR{pixel $\bf{p}'$ in view face}
        \STATE Find corresponding points ${\bf{p}}'_i$ for $\bf{p}'$ on edges $\mathcal{E}_i$
        \STATE Calculate 3D coordinate ${\bf{p}}_i$ for ${\bf{p}}'_i$ using pinhole camera model and spatial constraints
        \STATE Obtain $\bf{p}$ by interpolating between ${\bf{p}}_1$ and ${\bf{p}}_2$
    \ENDFOR
    \STATE Aggregate 3D coordinates into pointcloud $\mathcal{P}_{LUF}$
    \RETURN $\mathcal{P}_{LUF}$
}
\end{algorithmic}
\end{algorithm}

\subsection{Local Geometry Reconstruction}
\label{sec:local_reconstruction}
\begin{figure}[!t]
\centering
\includegraphics[width=\linewidth]{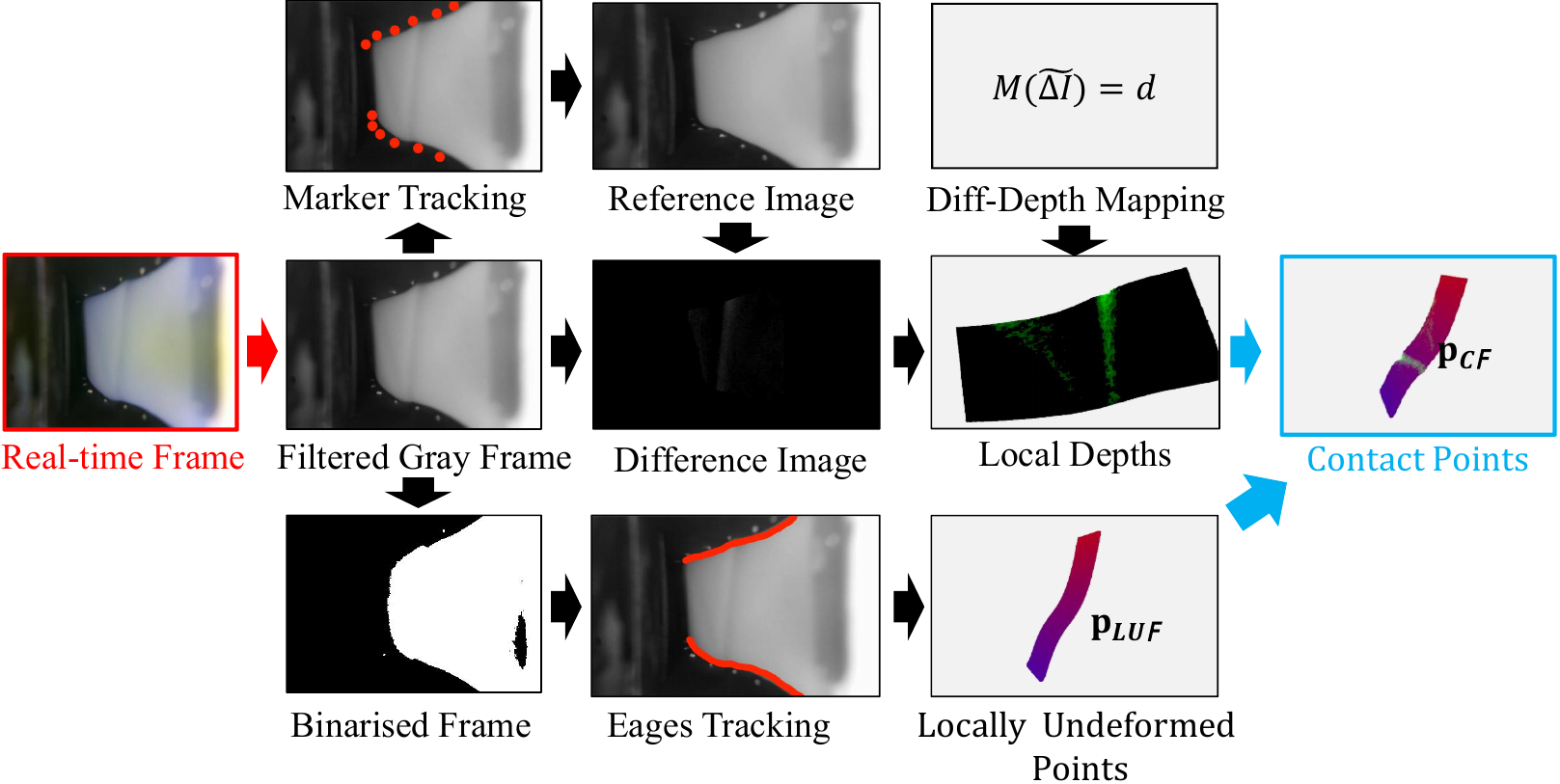}
\caption{{The pipeline of local geometry reconstruction. To handle the illumination changes caused by the global deformation, we dynamically retrieve the reference image from the video by using marker tracking, compute the brightness difference, and reconstruct the local depth. }}
\label{fig:local_pipeline}
\end{figure}

\begin{algorithm}[t]
\renewcommand{\algorithmicrequire}{\textbf{Input:}}
\renewcommand{\algorithmicensure}{\textbf{Output:}}
\caption{\textcolor{black}{Local geometry reconstruction}}\label{alg:local_geometry}
\begin{algorithmic}[1]
\REQUIRE \textcolor{black}{Grayscale image $I$ and locally undeformed face $\mathcal{P}_{LUF}$}
\ENSURE \textcolor{black}{Locally
deformed face $\mathcal{P}_{CF}$}
\textcolor{black}{
    \STATE Track marker to obtain marker coordinates ${\bf{m}}_k'$
    \FOR{frame $I_i$ in reference video}
        \STATE Calculate $d^{(i)}_{marker}$ using pre-computed marker coordinates ${\bf{m}}_k'^{(i)}$
    \ENDFOR
    \STATE Retrieve frame $I_i$ with the smallest $d^{(i)}_{marker}$ as the reference image $I_{ref}$
    \STATE Calculate normalized brightness difference $\widetilde{\Delta I}$
    \STATE Perform mapping $M()$ to get local depth $\mathcal{D}=M(\widetilde{\Delta I})$
    \STATE Obtain locally
deformed points $\mathbf{p}_{CF}$ by \newline ${\mathbf{p}}_{CF}={\mathbf{p}}_{LUF}-(0,0,d)^\top$ for each ${\mathbf{p}}_{LUF}$
    \STATE Aggregate 3D coordinates into pointcloud $\mathcal{P}_{CF}$
    \RETURN $\mathcal{P}_{CF}$
}
\end{algorithmic}
\end{algorithm}

In Sec.~\ref{sec:global_deformation}, we have localized all the points ${\bf{p}}_{\text{VF}}$ on the \textit{view face}.
In this section, we further localize all the locally deformed points {$\bf{p}_{\text{CF}}$} to reconstruct the contact geometry.
The pipeline of local geometry reconstruction is illustrated in Fig.~\ref{fig:local_pipeline} and Algorithm~\ref{alg:local_geometry}. 

{
As shown in Fig.~\ref{fig:principle}, 
The relative position between $\bf{p}_{\text{CF}}$ and the point on the \textit{locally undeformed face (LUF)} ${\bf{p}}_{\text{LUF}}$ is:
\begin{equation}
\label{eq:LUF_to_CF}
    {\bf{p}}_{\text{CF}} = {\bf{p}}_{\text{LUF}} - (0, 0, d)^\top
\end{equation}
where $d$ is the press depth. And the relative position between ${\bf{p}}_{\text{LUF}}$ and ${\bf{p}}_{\text{VF}}$ is:
\begin{equation}
\label{eq:VF_to_LUF}
    {\bf{p}}_{\text{LUF}} = {\bf{p}}_{\text{VF}} + (0, 0, t_0)^\top
\end{equation}
\begin{equation}
\label{eq:tv_to_t0}
t_{0} = \frac{t_{v}}{\text{cos }\theta}
\end{equation}
where $t_v$ is the thickness of the semi-transparent silicon and is known, $\theta$ is the incline angle at the point of \textit{view face}, which can be derived by polynomial fitting of the points on the edge.
Here, we simplify the computation by modeling the imaging process as parallel projection. Although a small error is introduced , it can be neglected due to the small incidence angle. 
}

{
According to the derivation above, the press depth $d$ is the sole unknown variable required to compute  $\bf{p}_{\text{CF}}$.
}
When \textit{contact face} is pressed and deforms locally, the contact region gets thinner, imaging thickness $t$ changes, and the brightness $I$ of a pixel within the region in the captured image changes accordingly, {which allows us to establish a mapping from \textit{$\Delta$I} to $d$.}
However, {different from DTact\cite{Dtact}}, the brightness of the pixel is also influenced by the distance to LEDs owing to the global deformation of our finger. In order to eliminate the influence, we establish a mapping \textit{M()} from the normalized brightness difference $\widetilde{\Delta I}$ to $d$: $M(\widetilde{\Delta I})=d$, and $\widetilde{\Delta I}$ is computed as:
\begin{equation}
\label{eq:delta I normalize}
\widetilde{\Delta I} = \frac{I_{ref}-I}{I_{ref}}
\end{equation}
where $I$ and $I_{ref}$ are the brightness in the reference and real-time image, respectively. 
{The mapping $M()$ is established through calibration, where} a small ball with a known radius is pressed onto the contact face {and} all locally deformed points ${\bf{p}}_{\text{D}}$ are considered {to be} on both the ball surface and contact face simultaneously.
{As shown in Fig.~\ref{fig:principle}, owing to the angled contact face relative to the camera's optical axis, the pressed region appears as an elliptical projection with a partially occluded circular boundary in the captured image. We fit a circle to maximally align with the circular contour of the dark region as the projection of the ball in the image and mark the center ${\bf{c}}'$ and an edge point ${\bf{p}}'$ of the circle on the image to localize the ball. }

As illustrated in Fig.~\ref{fig:principle}, the edge point ${\bf{p}}'$ corresponds to the point of tangency of the ray from the optical center to the ball {and} ${\bf{c}'}$ is corresponding to the ball center ${\bf{c}}$. 
We first compute the vectors ${\bf{p}_n}$ and ${\bf{c}_n}$, which represent the directions from $\text{O}_c$ to ${\bf{p}}$ and ${\bf{c}}$, through (\ref{pinhole imaging}) as:
\begin{equation}
\label{eq:unit_get}
\bf{p}_{n} = \bf{K}^{-1} \bf{p}' = \frac{\bf{p}}{\textit{z}}
\end{equation}
where $\bf{K}^{-1}$ is the inverse matrix of $\bf{K}$.
Then the 3D coordinates of ${\bf{c}}$ can be computed geometrically as:
\begin{equation}
\label{eq:c_get}
\frac{R}{\left \| \bf{c} \right \| } = \sqrt{1 - \left(\frac{\bf{p}_{n}\cdot \bf{c}_n}{\left \| \bf{p}_n \right \|
 \cdot\left \| \bf{c}_n \right \| } \right)^2} 
 \end{equation}
 \begin{equation}
     \bf{c}= \frac{\left \| \bf{c} \right \|}{\left \| \bf{c}_n \right \|} \cdot \bf{c}_n
 \end{equation}

With (\ref{pinhole imaging}), ${\bf{c}}$ and $R$, the points $\bf{p}_{\text{D}}$ could be computed. 
Having $\bf{p}_{\text{D}}$ and $\bf{p}_{\text{LUF}}$ computed from (\ref{eq:VF_to_LUF}), {the press depth is formulated as:}
\begin{equation}
\label{p to d}
    d_{\text{D}} =z_{\text{D}} - z_{\text{LUF}}
\end{equation}
where $z_{\text{D}}$ and $z_{\text{LUF}}$ are z-axis coordinates of $\bf{p}_{\text{D}}$ and $\bf{p}_{\text{LUF}}$. So far, we have derived $d_{\text{D}}$ and $\widetilde{\Delta I_D}$ from the dark region, which contains enough varying depths data, {allowing} a universal mapping $M()$ between any $d$ and $\widetilde{\Delta I}$ to be established {through polynomial fitting of} $d_{\text{D}}$ and $\widetilde{\Delta I}_\text{D}$, 
{thereby completing the calibration process.}

In addition, to ensure accurate {fitting} of the darker region {circle} contour in the calibration, we propose a semi-automatic calibration method. 
First, the program automatically detects an approximate contour of the region, whose size and position are further adjusted to improve the fitting accuracy.
{
Since a polynomial function is employed to represent the mapping function $M()$, it must map a zero input to a zero output, as, theoretically, no brightness difference should arise when the pressing depth is zero. Therefore, the constant term of $M()$ should be close to zero. Leveraging this property as a metric for assessing the quality of the fitting, the self-calibration method iteratively refines the size and position of the circle to search for an optimal $M()$ that best satisfies this criterion.
}

{During the sensing process,} a reference image without local contact is required to compute $\widetilde{\Delta I}$ and $d$.
{However, since the reference image dynamically changes due to the global deformation of the contact face, we cannot use a single reference image for all situations as in DTact [30]. Therefore, we use a video as a set of reference images to address the problem.}
Specifically, a video that contains a large amount of diverse deformation situations is recorded beforehand, in which the finger is not pressed on the contact face but rather pulled from the side or back faces, 
{ensuring sufficient} global deformation while the local deformation does not occur.

{
When using the finger, the markers doted on the side faces are tracked in the real-time captured image. Then we compute the marker distance $d_\text{marker}$ between the markers in the real-time captured image and those in each video frame:
\begin{equation}
\label{eq:marker_distance}
d^{(i)}_\text{marker}=\sum_{{\bf{m}}_k'\in \mathcal{M}}{\left \| {\bf{m}}_k'^{(i)}-{\bf{m}}_k' \right \|_2}
\end{equation}
where $\mathcal{M}$ is the set of markers, ${\bf{m}}_k'$ and ${\bf{m}}_k'^{(i)}$ are the image pixel coordinate of the $k$th marker in the real-time image and the $i$th reference image, respectively.
The frame with the smallest $d^{(i)}_\text{marker}$ is considered as the corresponding reference image and used for brightness difference computation in (\ref{eq:delta I normalize}). 
}

To compute $d^{(i)}_\text{marker}$, it is required to track the marker positions accurately and stably.
Learning-based approaches can learn to extract features that are robust to illumination changes, enabling more stable tracking of the markers.
In order to ensure the real-time contact detection capability, we use YOLO11~\cite{githubGitHubUltralyticsultralytics}, a lightweight fast object detection method, to detect the markers. We develop a fast algorithm to build the correspondences of the detected markers between frames for marker tracking, where the markers are divided into upper and lower groups according to the $y$-coordinates and further sorted from left to right to obtain the index $k$ shown in (\ref{eq:marker_distance}).
It is worth pointing out that some markers might get occluded when the finger encounters large deformation. Under these circumstances, the motion of occluded markers is estimated by using the adjacent visible markers.

\begin{figure}[!t]
\centering
\includegraphics[width=0.7\linewidth]{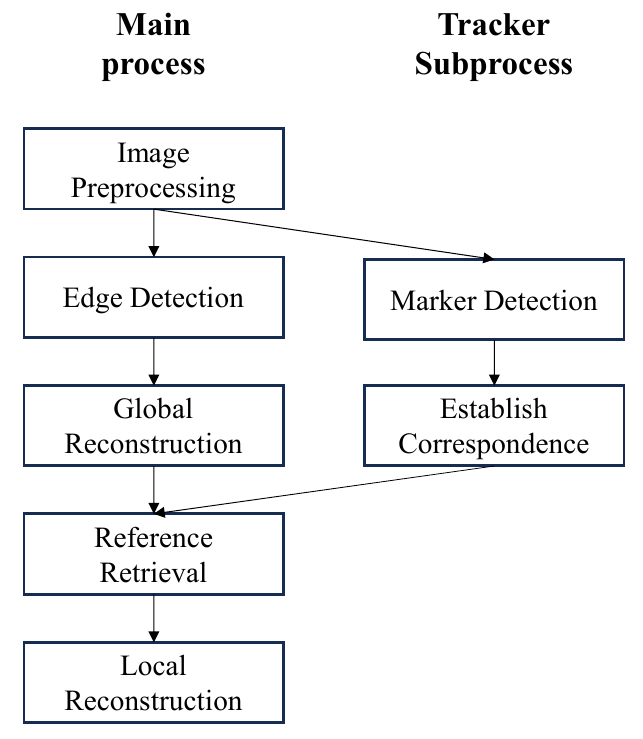}
\caption{{The multiprocessing workflow: the global reconstruction is performed by the main process, while the marker tracking is handled by the tracker subprocess  concurrently to reduce the computation time.}}
\label{fig:multi_process}
\end{figure}

To further minimize the computation time and latency, we utilize multiprocessing techniques, as illustrated in Fig.~\ref{fig:multi_process}.
Once an image is captured by the camera, it first undergoes preprocessing, including resizing, flipping, and conversion to grayscale. The preprocessed image is then stored in shared memory, allowing direct access by the marker tracking and detection subprocess. The edge detection and global reconstruction is handled by the main process simultaneously. Once the subprocess outputs the indexed marker pixel coordinates, which are also saved to shared memory, the main process performs reference image retrieval, and local geometry reconstruction.

\subsection{Omni-Directional Contact Detection}
\label{sec:contact_detection}
Unexpected contacts may occur during robot grasping and manipulation, which 
{poses significant risks of task failure and mechanical damage.}
Therefore, it is critical for the gripper to detect the contact from all the directions. 

In this section we describe our method for omni-directional contact detection. We first track the marker positions in the real-time image and compute the offset to their initial positions: 
\begin{equation}
    {\bf{o}'} = \sum_{{\bf{m}}'\in \mathcal{M}}{ ({{\bf{m}}' - {\bf{m}}'^*})}
\end{equation}
where ${\bf{o}'} \in \mathbb{R}^2$ is the global offset and ${\bf{m}}'^*$ is the marker's initial position.

A contact is detected when $\left \|{\bf{o}}' \right \| > \varepsilon$, where $\varepsilon$ is a predefined threshold. As we align the finger's width direction to the $y$-axis of the camera, the contact direction in the 3D space is determined as:
\begin{equation}
    {\bf{o}}  = \frac{1}{\left \|{\bf{o}}' \right \|_2}({\bf{o}}'_u, {\bf{o}}'_v, 0)^\top
\label{contact direction}
\end{equation}
where ${\bf{o}}\in \mathbb{R}^3$ is a unit vector representing the force's direction. We compute the discrepancy between ${\bf{o}}$ and the normal vectors of all the four faces using inner product, and the face with the smallest discrepancy is determined as forced face. 

\begin{figure}[!t]
\centering
\includegraphics[width=0.95\linewidth]{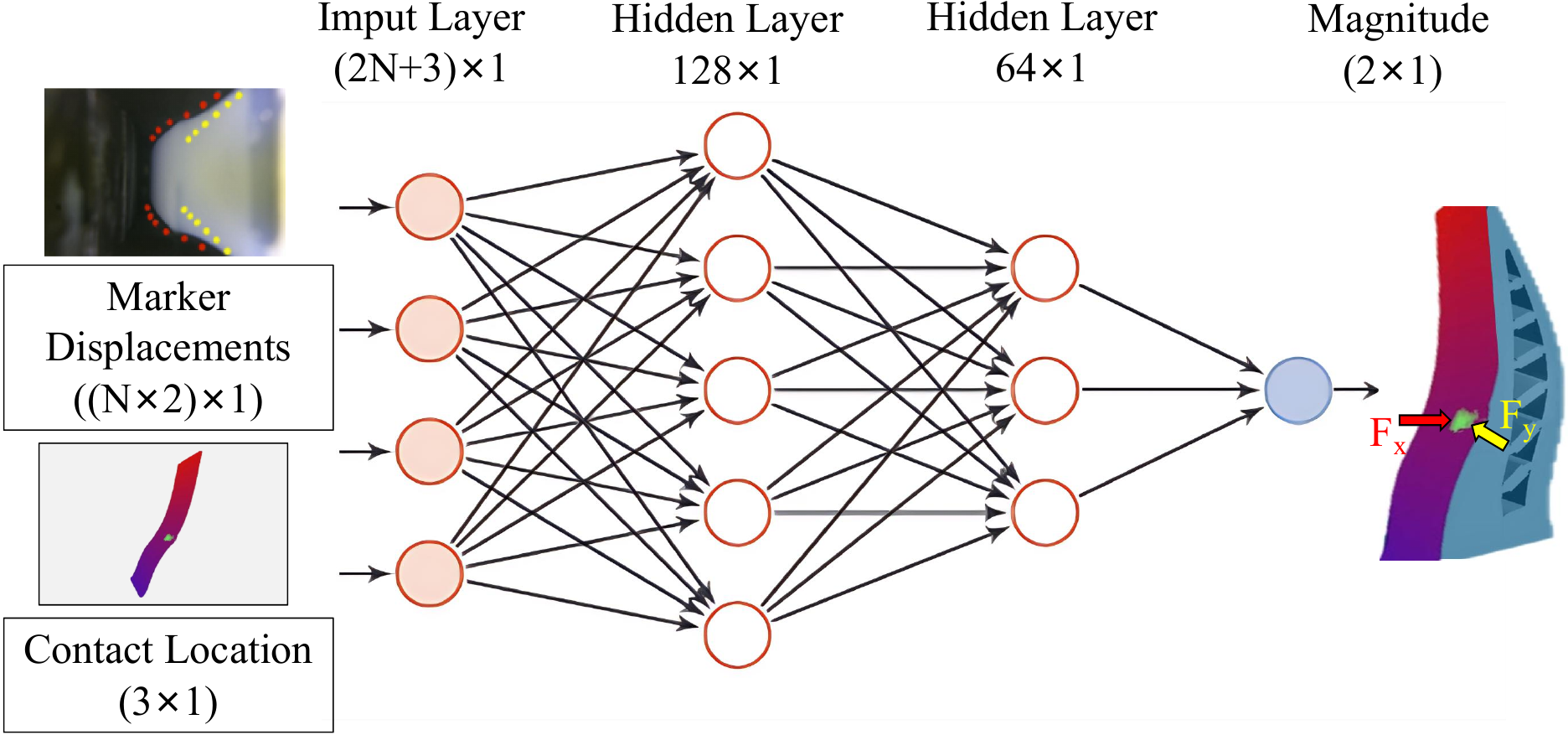}
\caption{{Neural network architecture for force magnitude prediction}}
\label{fig:network}
\end{figure}

{
Furthermore, perception of force magnitude at the contact face is crucial for manipulation, especially for delicate objects. To address this, a neural network is developed to predict the contact force magnitude using marker displacements and contact location. As shown in Fig.~\ref{fig:network}, the network’s input consists of the marker displacements and the 3D coordinate of the contact point, and the output is the force magnitudes along the $x$- and $y$-axes. We employ a multilayer perceptron (MLP) architecture with two hidden layers, with ReLU as activation functions.
}

\section{Experiments}
\label{sec:experiments}

In this section we conduct extensive experiments to verify the effectiveness of our finger and gripper. We first carry out contact {localization}, pose estimation, force detection for the finger. Then we perform object grasping and pose adjustment for the gripper. The visualized point clouds reconstructed by the sensor are also presented.

\begin{figure}[!t]
    \centering
    \subfloat[]{
        \includegraphics[width=\linewidth]{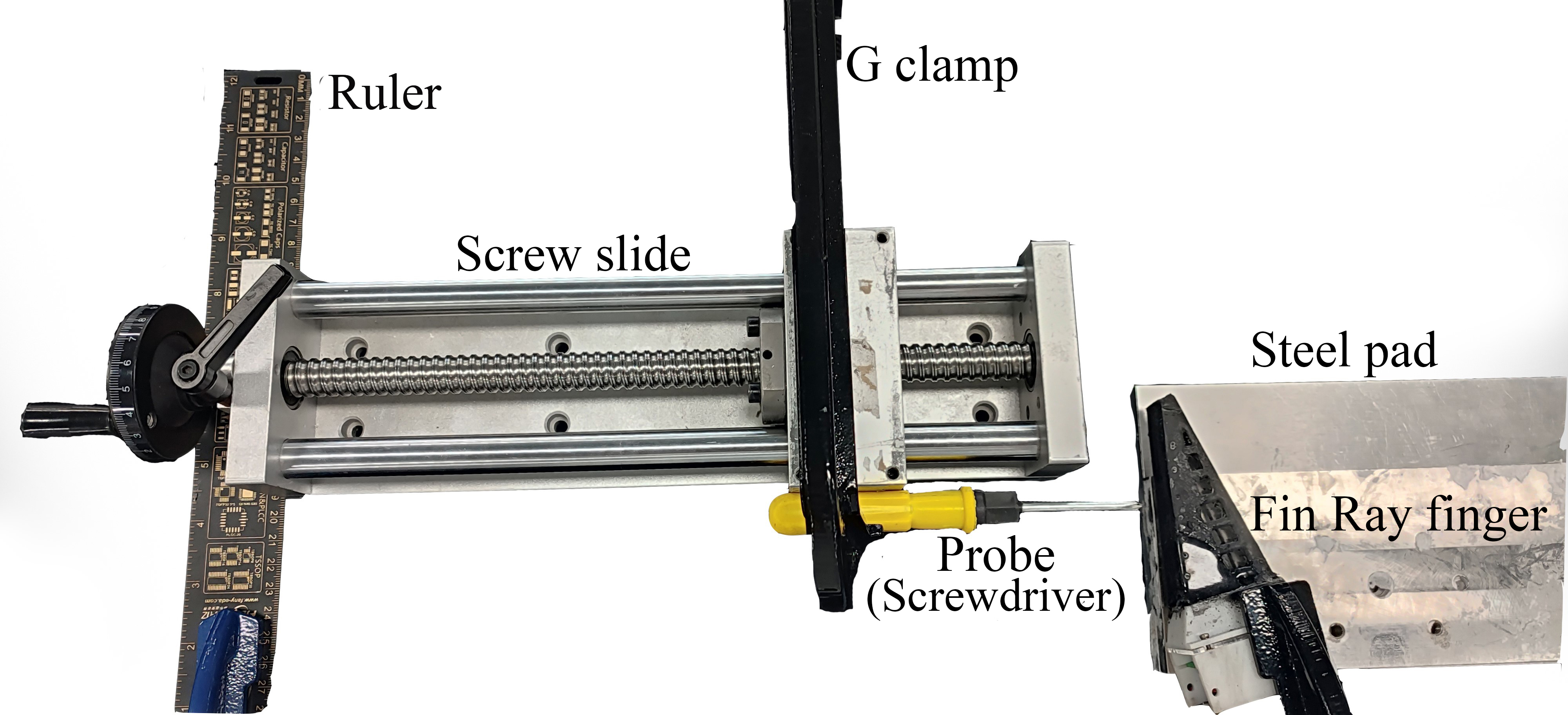}
        \label{fig:setup}}
    \hfill
    \subfloat[]{
        \includegraphics[width=0.35\linewidth, angle=90]
        {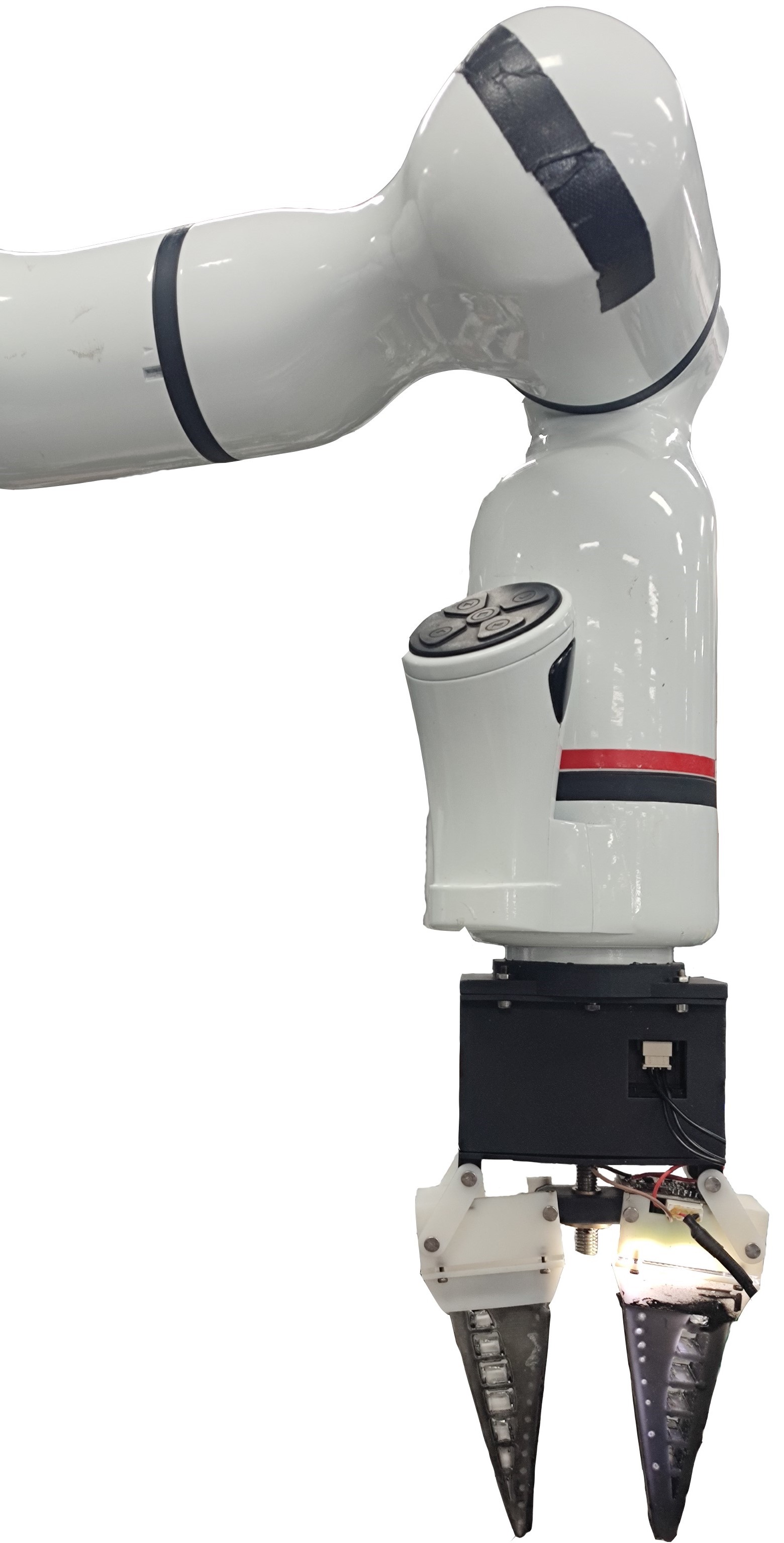}
        \label{fig:xmate3}}
\caption{Experimental setups. (a) Basic setup for sensor experiments, a screwdriver is used as the probe for contact {localization} experiment. (b) The gripper is mounted on a ROKAE xMate3-pro robotic arm.}
\label{fig:setup_and_arm}
\end{figure}

The experimental setups for the finger and gripper are shown in Fig. \ref{fig:setup_and_arm}. We use a ROKAE xMate3-pro robotic arm for the gripper. For the experiments of the finger, a probe is required to move towards the fixed finger. The setup consists of a screw slide using G-clamp to hold the probe to feed, a ruler fixed with G-clamp to measure distance, a few steel pads to adjust the heights of the finger, and the finger held by another G-clamp. The probe of the setup differs in each experiment.
\subsection{Contact {Localization}}
\label{sec:Contact Location}
\begin{figure}
    \centering
    \includegraphics[width=\linewidth]{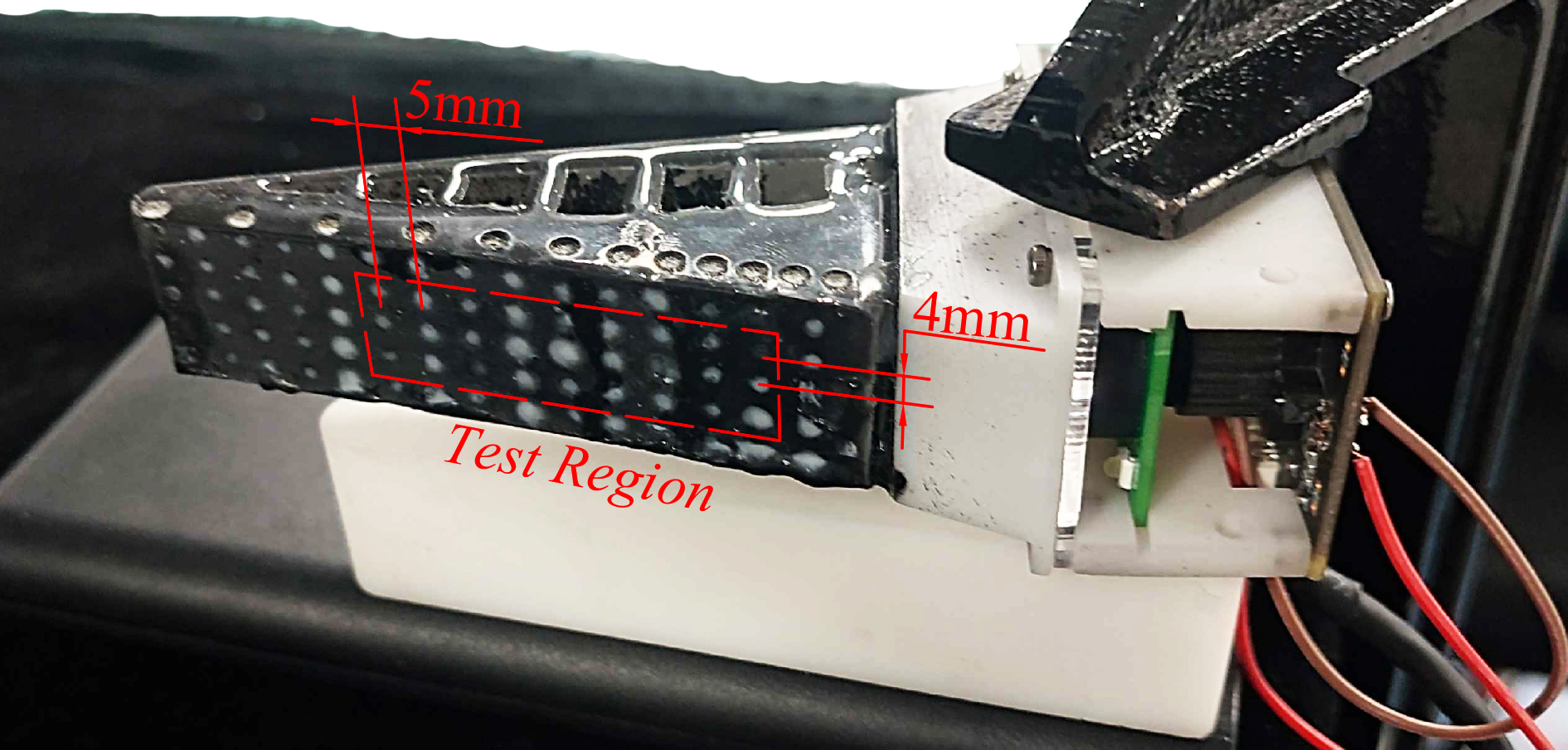}
    \caption{The dotted array on finger for contact {localization} experiment.}
    \label{fig:local_array}
\end{figure}

In this experiment, we measure the sensor {localization} precision for contacts on the contact face. We dot a marker array on the contact face (only for experiment) and each marker is poked {by} a screwdriver with a tip radius of 2 mm, as shown in Fig. \ref{fig:setup}. The poking positions from the sensor are compared with the {Groundtruth (GT)} marker positions. 
The marker array we dot has 9 rows with spacing of 5 mm and 3 columns with spacing of 4 mm, as shown in Fig.~\ref{fig:local_array}. 
We select a marker at the array center as the origin, compute its distances to all of the tested markers, and compare them with results from the sensor. 

\label{sec:force detection}

\begin{figure}
    \centering
    \includegraphics[width=0.6\linewidth]{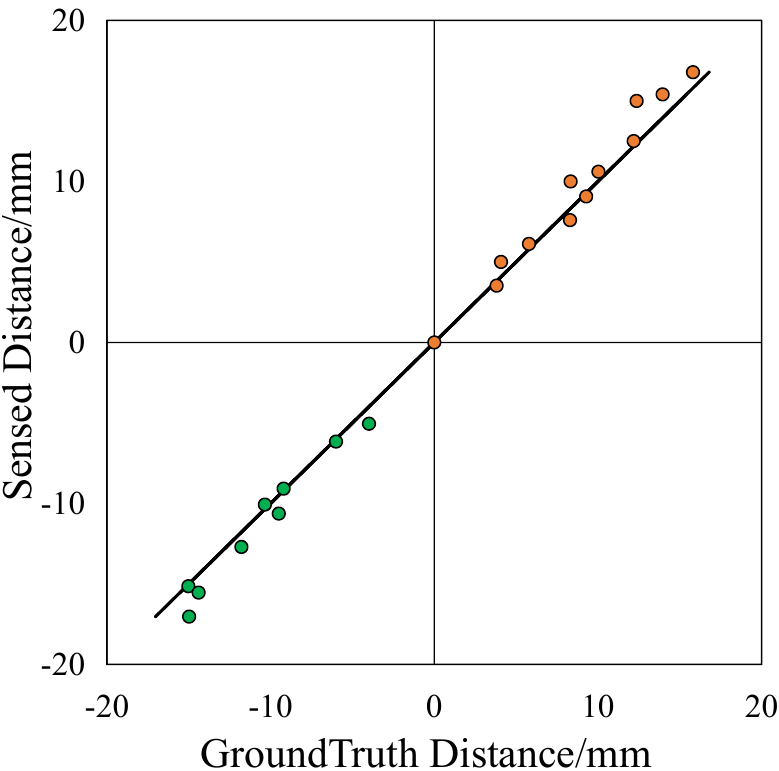}
    \caption{Contact location experiment result. Our method is able to accurately local the contact position in a range of -20 mm to 20 mm.}
    \label{fig:local_result}
\end{figure}

Fig. \ref{fig:local_result} shows the experiment results.
{
 Within a range of $\pm$20 mm, the mean absolute error (MAE) of contact localization is 0.814 mm, demonstrating that our finger can precisely localize the contact position. However, the MAE increases to 2.93 mm for positions near the fingertip. This discrepancy arises due to a reduction in illumination intensity, which results in greater imaging noise and reduces the accuracy of $\widetilde{\Delta I}$.
}

\begin{figure}[t]
\centering  
    \subfloat[]{
        \includegraphics[width=0.7\linewidth]{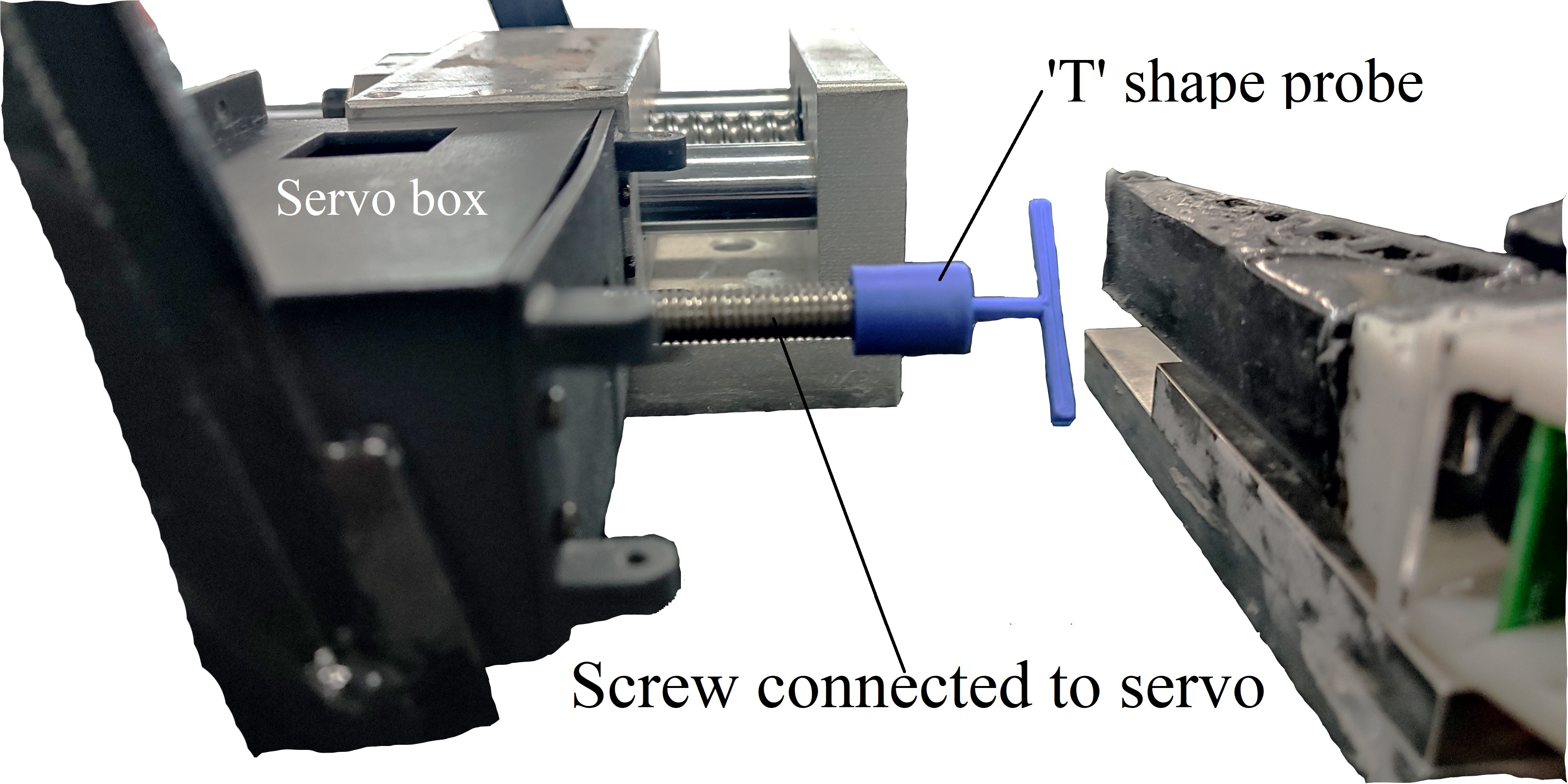}
        \label{fig:pos_experiment}}
    \hfill
    \subfloat[]{
    \includegraphics[width=0.7\linewidth]{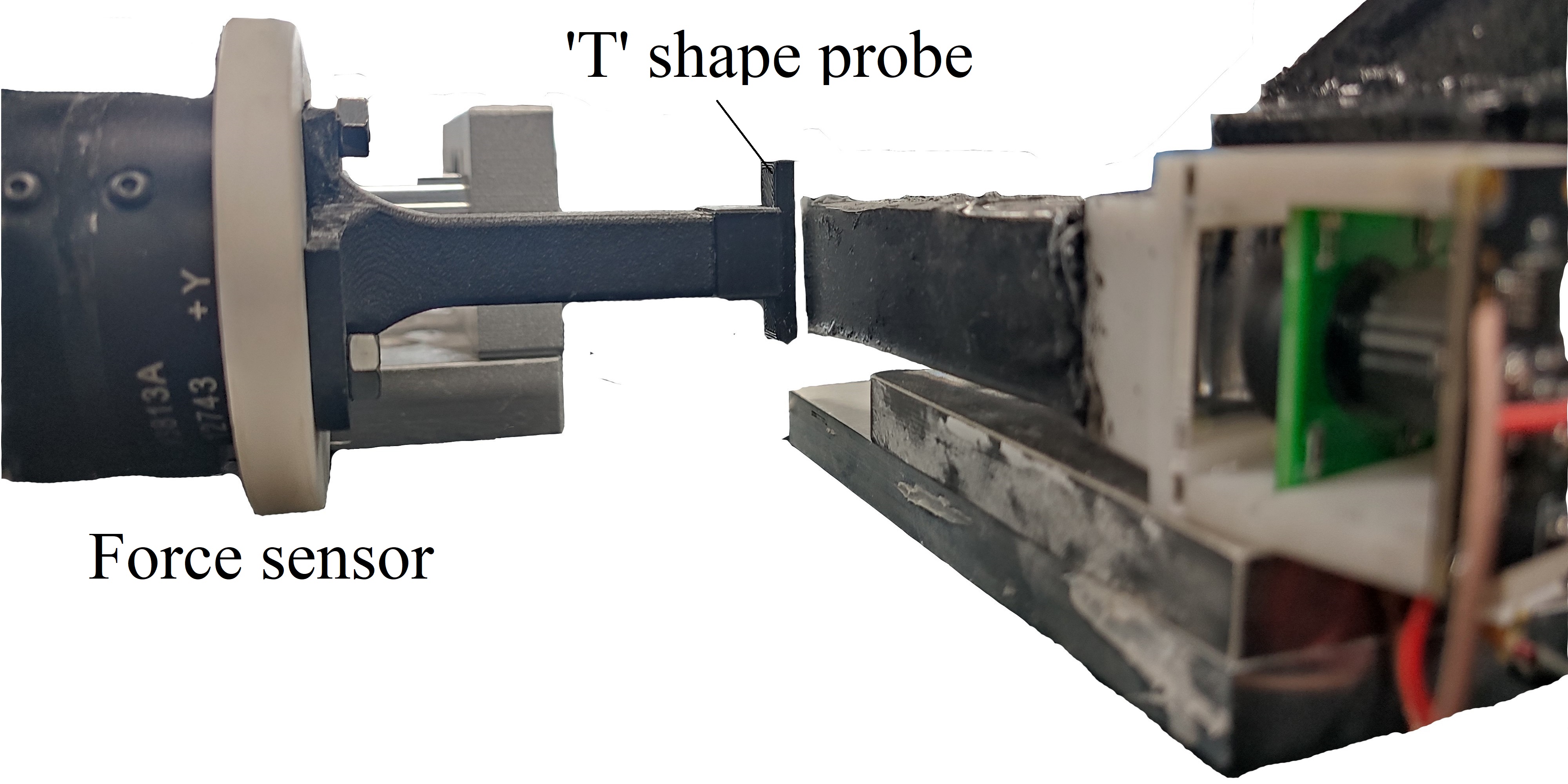}
    \label{fig:face_experiment}}
\caption{Other sensing experiment probes. (a)“T” shape probe in pose estimation experiment. (b)“T” shape probe on force sensor in force detection experiment.}
\label{fig:sensing_experiment}
\end{figure}

\subsection{Pose Estimation}
\label{sec:pose Estimation}
\begin{figure}
    \centering
    \includegraphics[width=0.6\linewidth]{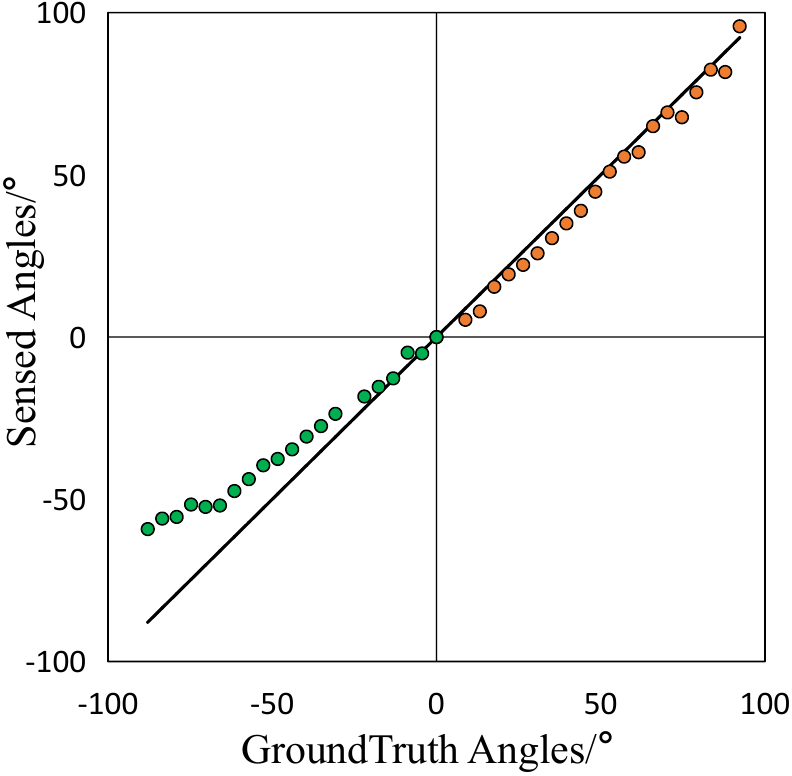}
    \caption{Pose estimation experiment result. Our method is able to accurately estimate the pose from the contact in a range of -100$\degree$ to 100$\degree$.}
    \label{fig:pos_result}
\end{figure}
{
The pose estimation refers to determining the pose of the grasped object pressed against the contact surface relative to the camera coordinate system. The experimental setup is illustrated in Fig.\ref{fig:pos_experiment}. A sleeve with a slot and a ``T''-shaped insertion probe is designed, which is screwed onto the screw connected to the servo, and the ``T''-shaped probe is inserted into the slot on the sleeve. The servo is mounted within a servo box and secured to the slide using a G-clamp. The probe is then rotated to a series of angles by the servo and pressed onto the finger.
}

We first reconstruct the point cloud of the contact region, and compute the angle using Principal Component Analysis (PCA). The GT angles are acquired from the servo. Fig.~\ref{fig:pos_result} shows the comparison results.
{The probe is rotated in two directions, which exhibits different accuracy. The MAE of the sensed rotated angles is 3.22$\degree$ in the clockwise direction, while it increases to 12.059$\degree$ in the counterclockwise direction. The reason for the discrepancy is that the ``T''-shaped probe is not accurately centered on the middle of the finger face, resulting in different contact lengths. When rotated counterclockwise, the probe predominantly contacts regions close to the fingertip, where the localization error is larger} 

\subsection{Contact Detection}
\label{sec:force detection}
\begin{figure}
    \centering
    \includegraphics[width=\linewidth]{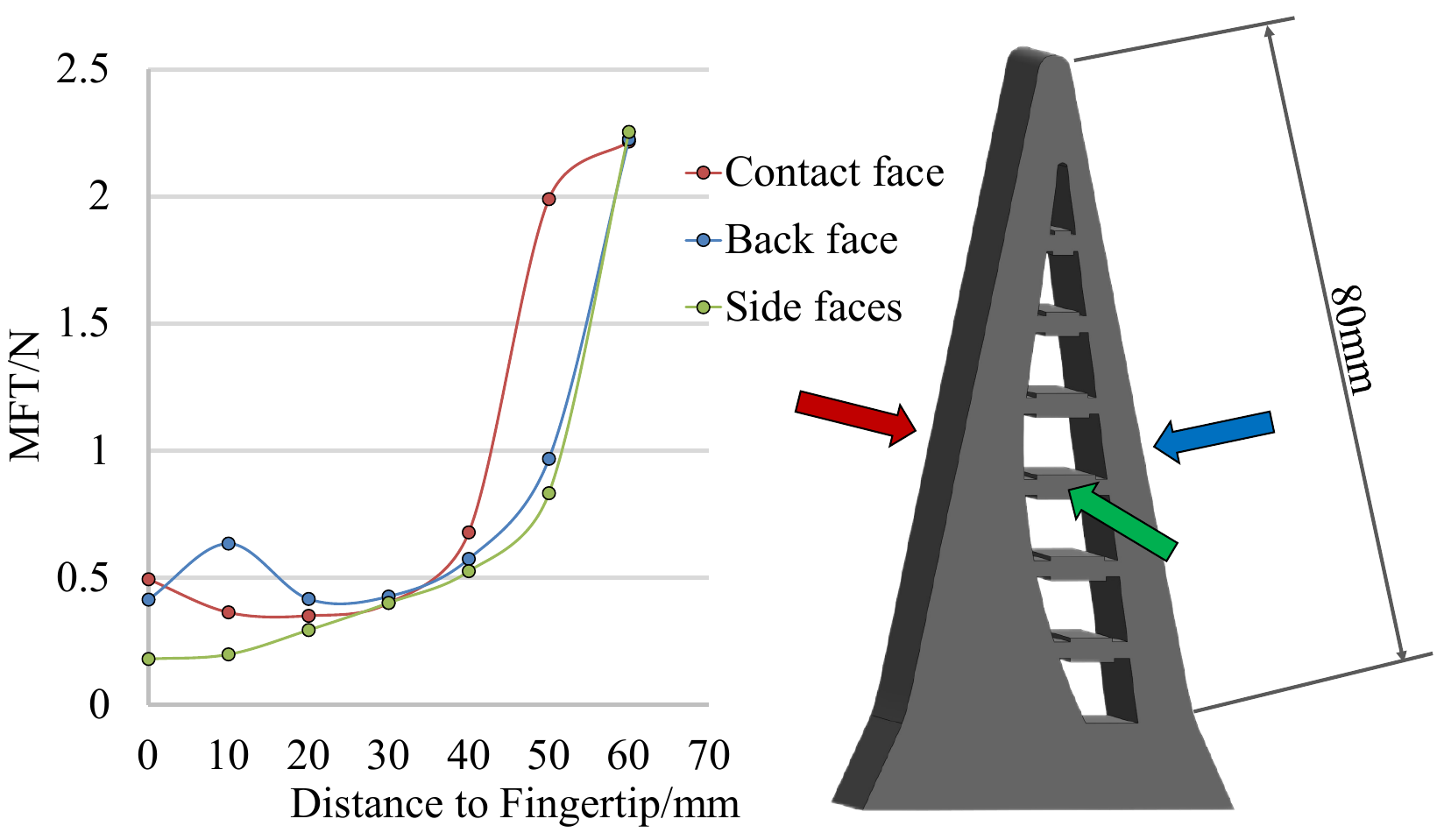}
    \caption{Contact detection experiment result, data and their corresponding faces are annotated with different colors. We average the data of the two side faces in the plot.}
    \label{fig:force_result}
\end{figure}

Fig.~\ref{fig:face_experiment} illustrates the experimental setup, where a ``T''- shaped probe is used to push all the four faces of the finger at a series of heights. An SRI M3813A Force/Torque sensor is used to measure the GT contact force. 
We measure the minimum force threshold (MFT) of the gripper in different directions. The MFT is defined as the minimum force that the sensor can detect the deformation and determine the direction, which reflects the sensor's sensitivity.
Fig.~\ref{fig:force_result} shows the experiment result, where the MFTs of the two side faces are averaged. {
The finger exhibits greater sensitivity to contacts at the side faces, which can be attributed to the finger's Fin Ray structure.
The MFT is lowest close the fingertip and increases toward the finger root due to the thicker wall design.
{Nevertheless}, regions beyond 20 mm from the finger root exhibit MFTs below 2 N, indicating that the majority of the finger can sensitively detect contacts from all directions. Since unexpected contacts during grasping and manipulation rarely occur near the root, this contact detection capability enhances the safety of robotic manipulation by enabling the robot to detect omni-directional collisions.
}

\begin{figure}
    \centering
    \subfloat[]{\includegraphics[width=0.49\linewidth]{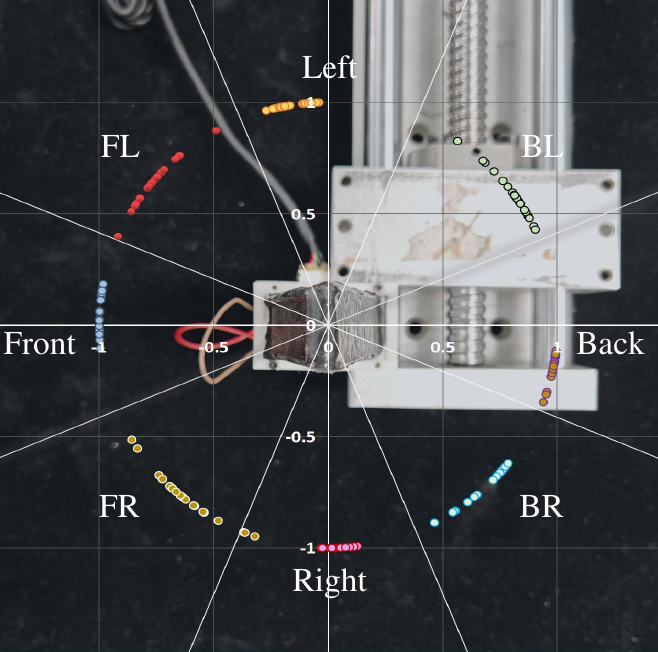}
    \label{fig:direction}}
    \subfloat[]{\includegraphics[width=0.49\linewidth]{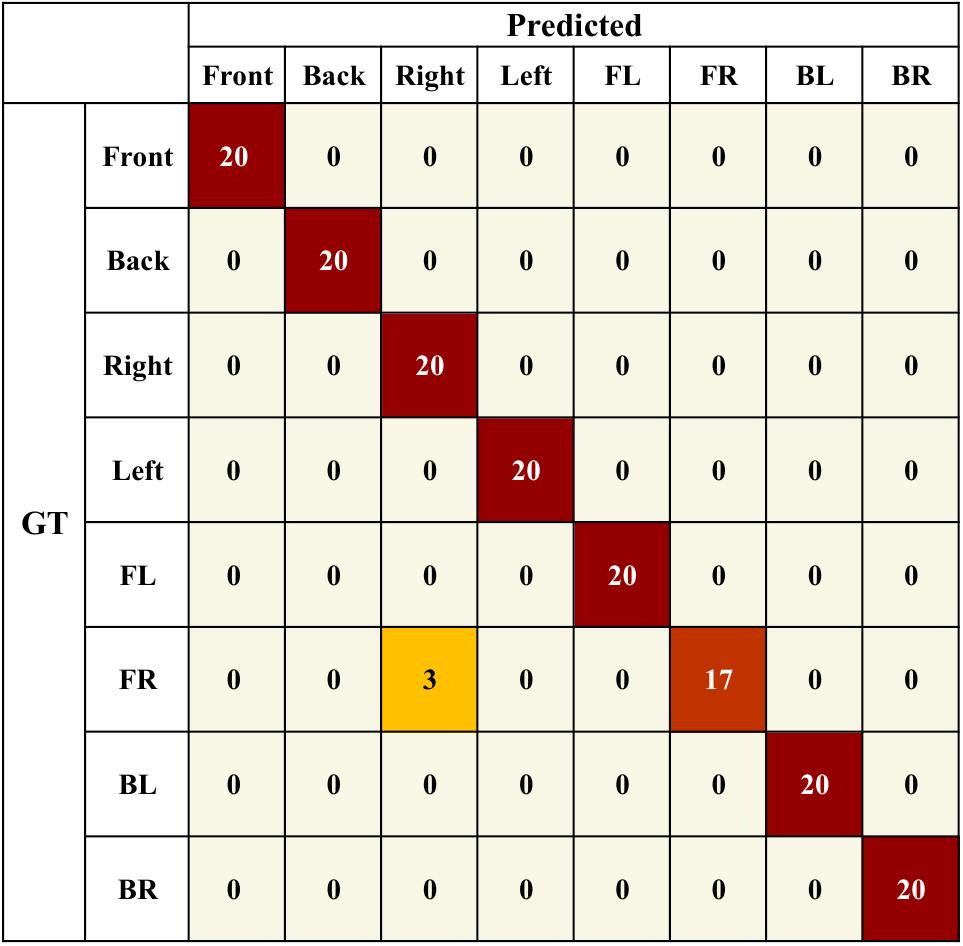}
    \label{fig:matrix}}
    \caption{Contact direction experiment result. (a) Measured contact direction; (b) Confusion matrix of contact direction classification.}
    \label{scatter and matrix}
\end{figure}

To validate our method's ability to measure contact direction, we fix the the finger upright to the screw slide, divide the $x-y$ plane into 8 regions and manually apply forces randomly on various heights and directions to the finger 20 times per region.
Fig. \ref{fig:direction} visualizes the 8$\times$20 contact directions computed using (\ref{contact direction}), demonstrating that our method can effectively measure contact from various directions.
To further evaluate the accuracy, we compare the predicted region index with the ground truth. As shown in Fig.~\ref{fig:matrix}, our method achieves a classification accuracy of 98.1\%, validating the contact detection accuracy of our finger and sensing method.

\begin{figure}
    \centering
    \subfloat[]{\includegraphics[width=0.9\linewidth]{experiment_fig/force_setup.pdf}
    \label{fig:magnitude_setup}}
    \hfill
    \subfloat[]{
        \includegraphics[width=0.45\linewidth]{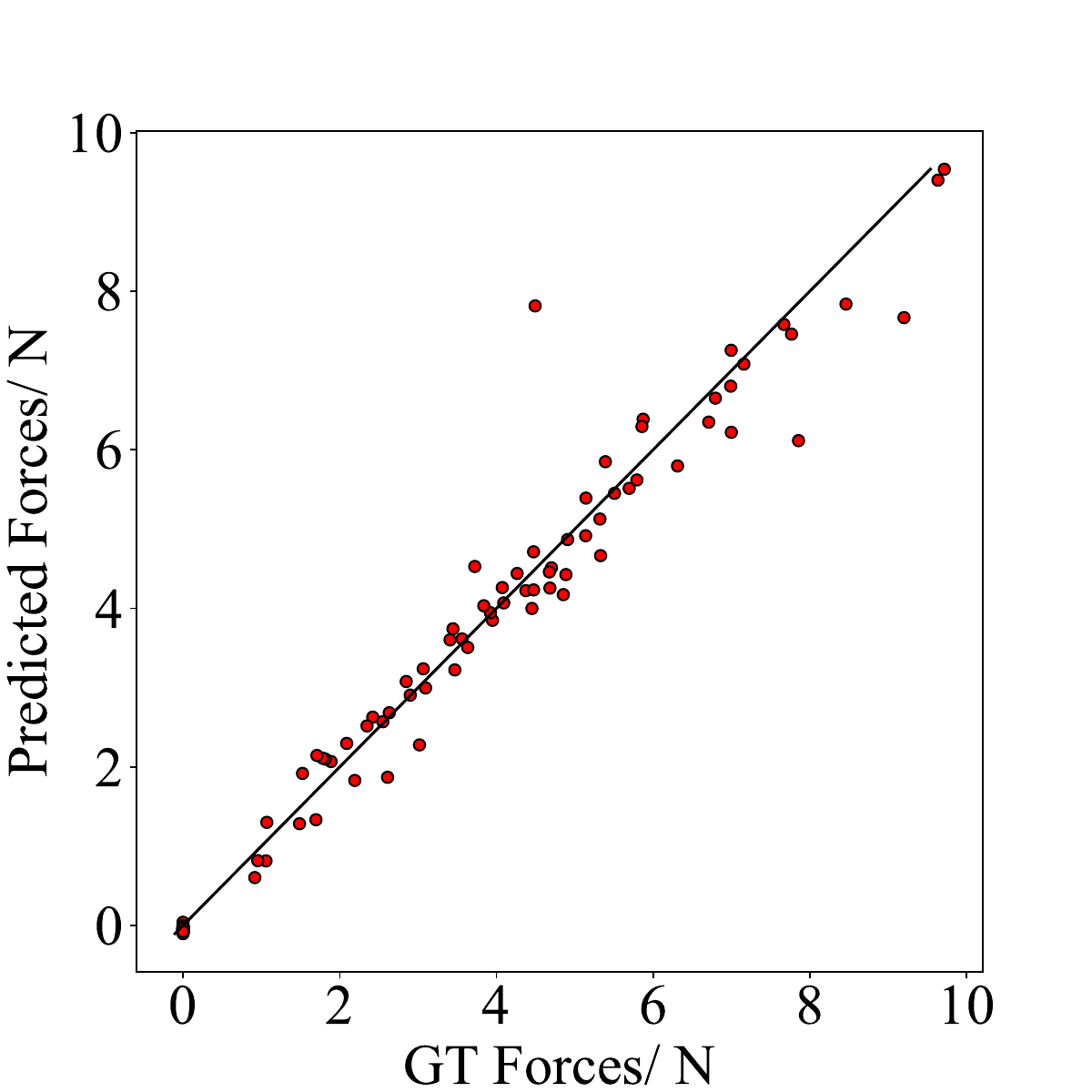}
        \label{fig:magnitude_x}}
    \subfloat[]{\includegraphics[width=0.45\linewidth]{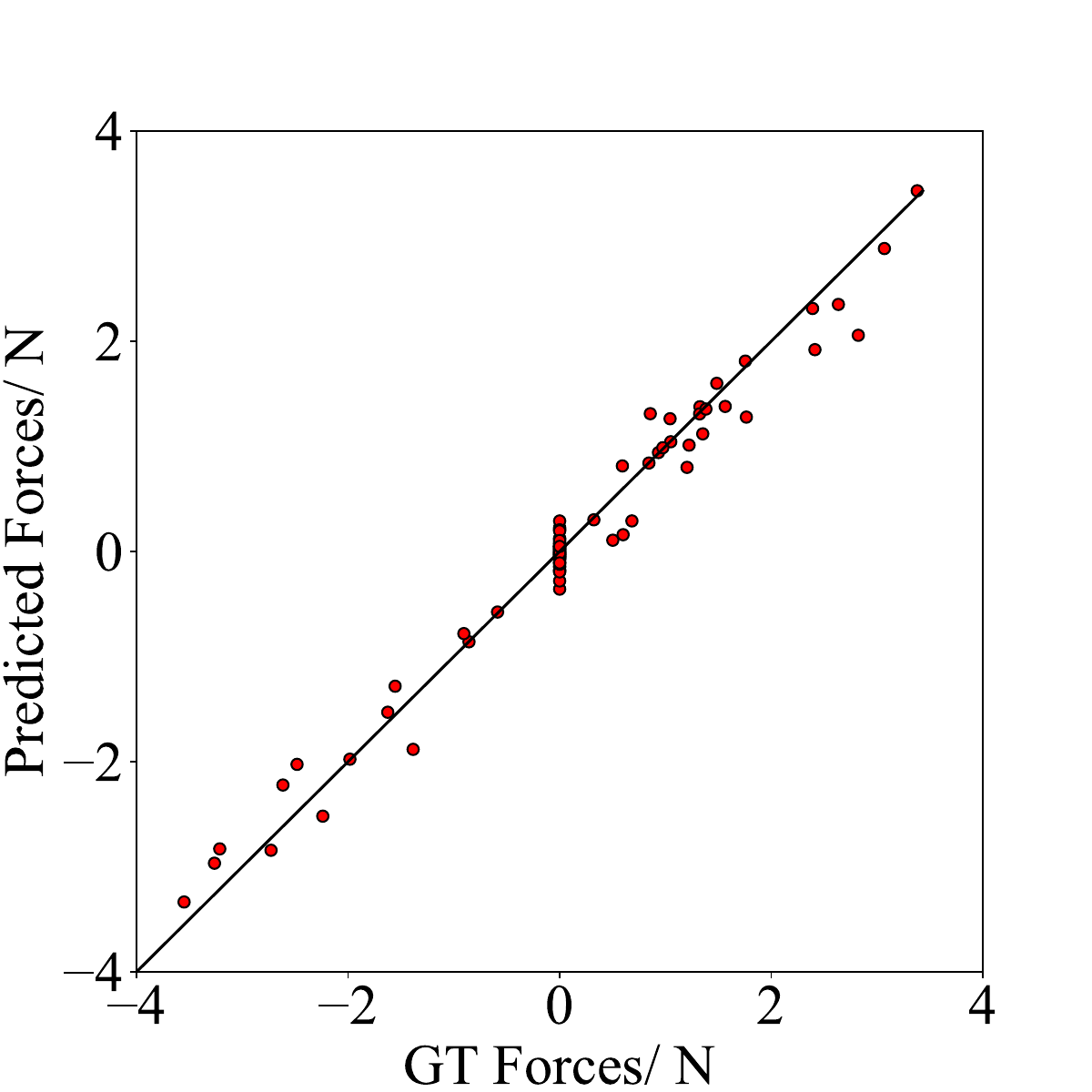}
    \label{fig:magnitude_y}}
    {
    \caption{Force magnitude prediction test set up and result. (a) Experiment setup; (b) Force Magnitude in x direction result; (c) Force Magnitude in y direction result}
    \label{fig:magnitude_setup_and_result}
    }
\end{figure}

{
We further evaluate the force magnitude prediction accuracy. Experiment setup is shown in Fig.~\ref{fig:magnitude_setup}, where the finger is fixed onto the desk, a robot arm is used to apply forces onto the contact face at varying locations and in different $x$-$y$ directions, and a lamp is employed to adjust the ambient light intensity.
The finger used in this experiment is dotted 6 markers on each side face. Thus, the dimension of the input data is $6 \times 2 \times 2 + 3 = 27$. 
A total of 484 sample points are collected and randomly divided into training and testing sets in an 8:2 ratio. The force ranges are 0 to 10 N along the $x$-axis and -4 N to 4 N along the $y$-axis. Experimental results are shown in Fig.\ref{fig:magnitude_x} and \ref{fig:magnitude_y}. The mean prediction error is 0.244 N for the $x$-axis and 0.162N for the $y$-axis, demonstrating the sensor’s capability for precise force magnitude prediction.
}

\subsection{Local Geometry Reconstruction}
\label{sec:local reconstruction result}
\begin{figure}[]
\centering 
\includegraphics[width=0.9\linewidth]{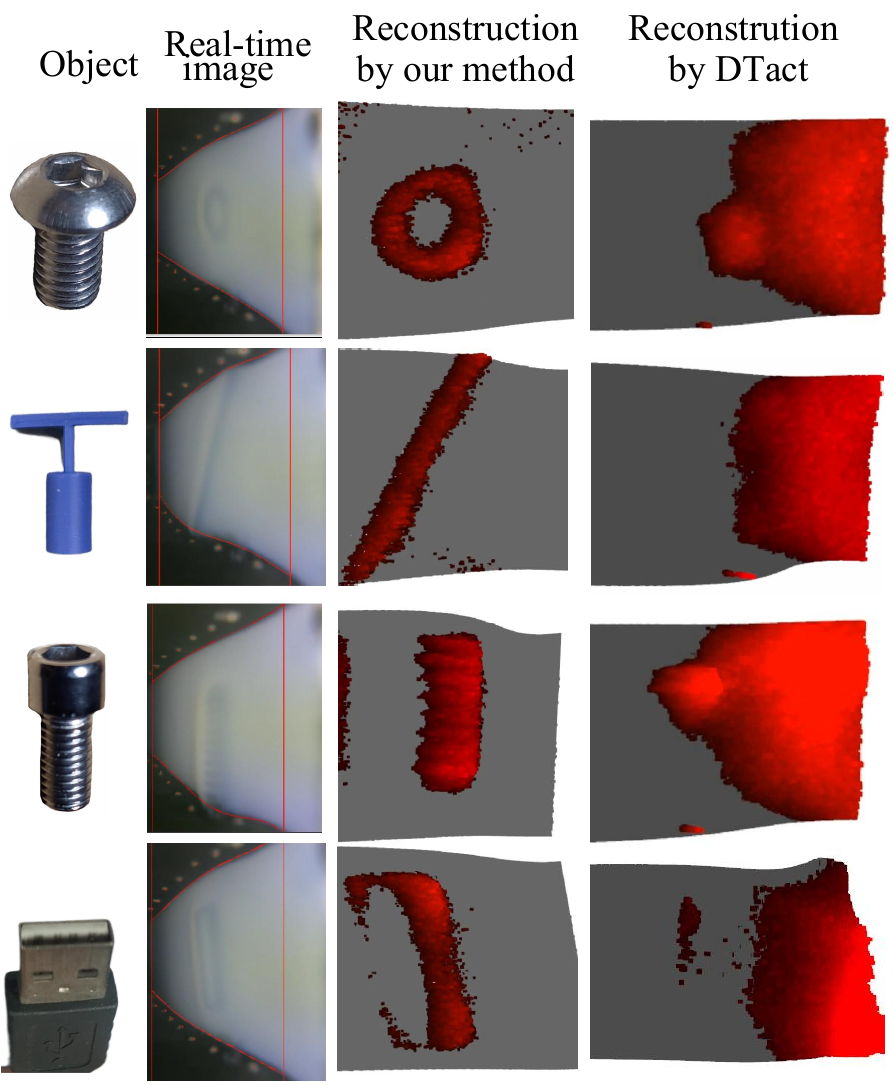}
\caption{{Contact geometry reconstruction comparison: the reconstruction results by our method are more clear and accurate than those by DTact~\cite{Dtact}. }}
\label{fig:reconstruct_images}
\end{figure}

{Fig.~\ref{fig:reconstruct_images} compares the local geometry reconstruction results by our method and that in DTact~\cite{Dtact}. 
Our finger is able to reconstruct the detailed contact geometry, capturing intricate features such as the threads of an M8 screw.
Since the single-reference method fails to account for the brightness changes caused by global deformation, the reconstruction results contain severe noise in non-contact regions. In contrast, our method mitigates such artifacts, producing clean, noise-free reconstructions for the contact geometry.}

\begin{figure}
    \centering
    \subfloat[]{\includegraphics[width=0.48\linewidth]{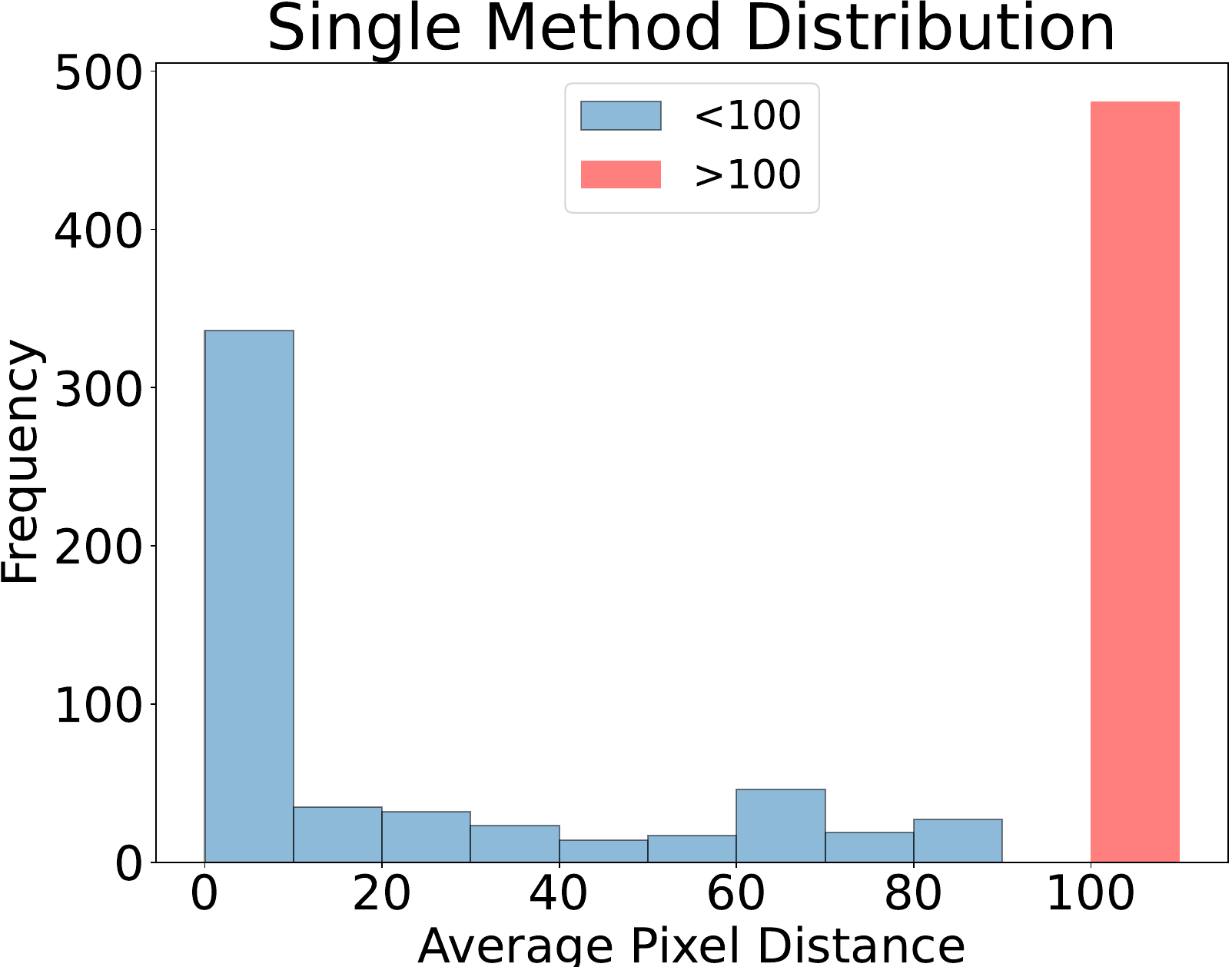}
    \label{fig:single_error}}
    \subfloat[]{\includegraphics[width=0.48\linewidth]{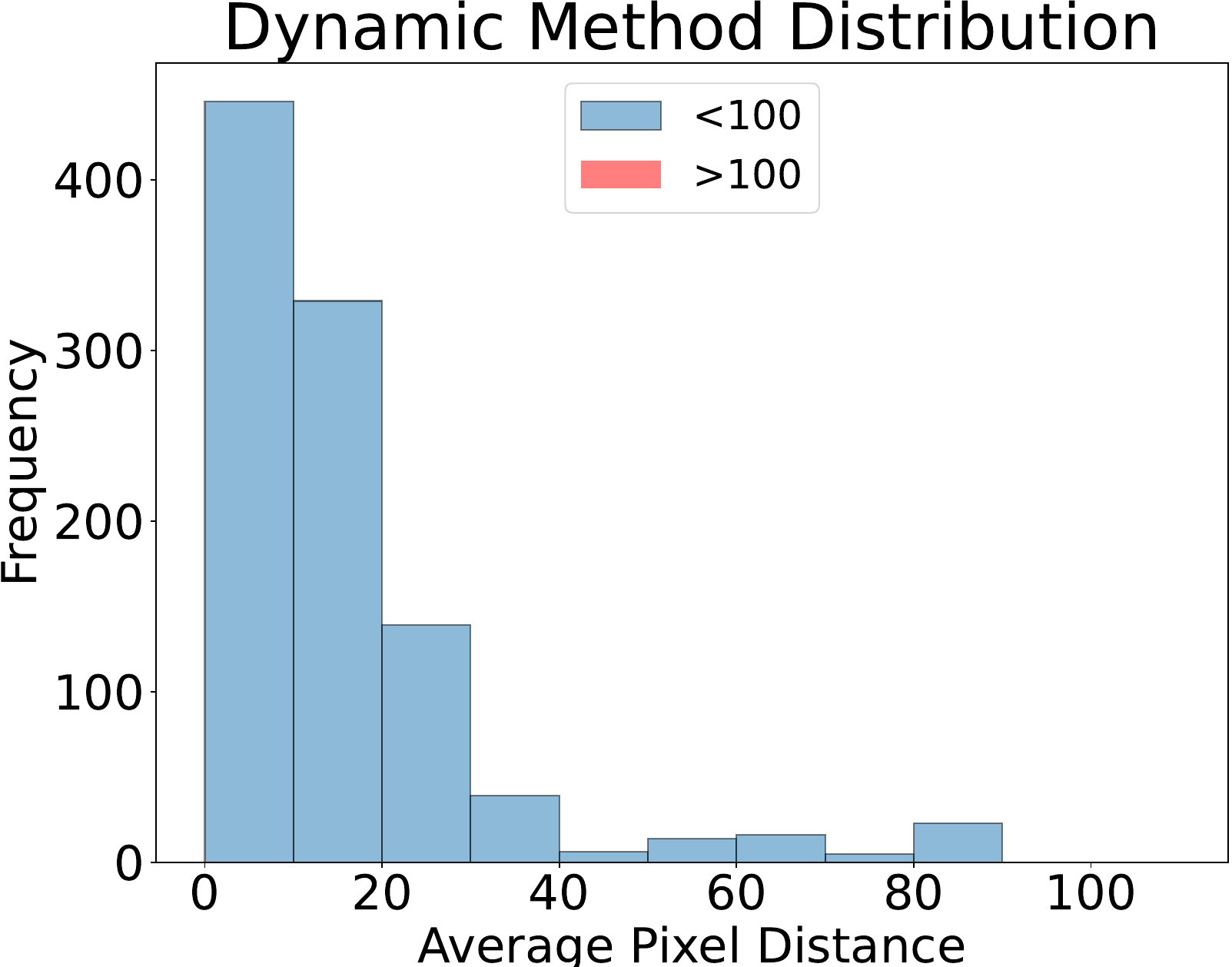}
    \label{fig:dynamic_error}}
    \caption{Mean pixel error frequency statistics in different referring methods. (a) Mean pixel distance distribution referred from single image; (b) Mean pixel distance distribution referred by using our proposed dynamic referencing method.}
    \label{referring error}
\end{figure}

{The length of the reference video utilized in our experiments for two fingers are 52s and 55s, respectively, at a FPS of 30.
In order to demonstrate the alignment accuracy between the real-time images and their corresponding reference images extracted from the video, we calculate the average pixel distance error of the markers across the images.
We record a video consisting of 1050 frames as ``real-time'' images where the finger gradually contacts the object and then separates. We compute the 1050 distances between the real-time images and the dynamically retrieved reference images using (\ref{eq:marker_distance}), which are further divided by the number of markers as average distances. Fig.~\ref{referring error} visualizes the frequency distribution of the distances. The statistics show that 91.0 \% of the mean distances are lower than 40 pixels in dynamic reference method, while in single reference method the value is reduced to 40.6 \%. The result indicates that our method is able to dynamically retrieve the corresponding reference image from the video that has similar global deformation to the target image.
}

\begin{table}[]
    \centering
    {
    \caption{Sensor performance under different ambient light intensities}
    \label{tab:ambient_light}
    \begin{tabular}{c|ccc}
    \toprule
      Ambient light intensity   &  300 Lux & 1100 Lux & 2500 Lux \\
      \midrule
       Localization error  & 1.872 mm & 2.145 mm & 1.923 mm \\
    Force magnitude error & 0.244 N & 0.235 N & 0.218 N\\
       \bottomrule
           
    \end{tabular}
    }
\end{table}

{
\subsection{Impact of Ambient Light Conditions}
In this section, we investigate the effects of ambient light intensity on contact localization, force magnitude estimation and geometry reconstruction. As shown in Fig.~\ref{fig:magnitude_setup}, we use a lamp to adjust the ambient intensity to three levels: 300 Lux, 1100 Lux and 2500 Lux. Table~\ref{tab:ambient_light} shows the experimental results. As can be seen, the ambient light intensity has a subtle impact on the sensor’s performance. The reason is that the finger is fully painted black, which effectively blocks ambient illumination and minimizes the influence.
\begin{figure}[]
    \centering
    \includegraphics[width=0.95\linewidth]{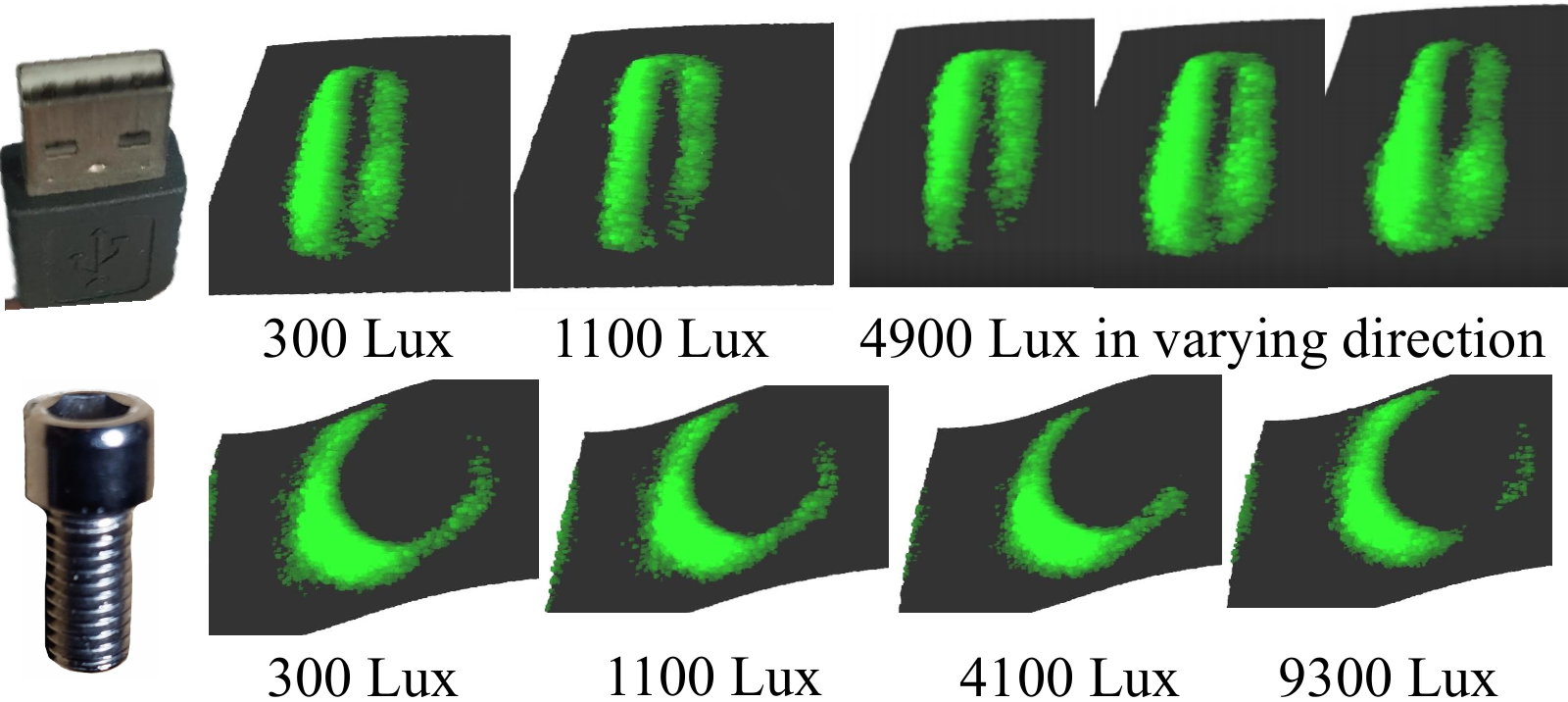}
    \hfill
  {\caption{Local reconstruction results under varying light intensity, where the calibration is performed at 300 Lux. The objects include a USB stick and an M12 hexagon socket round head screw.}
    \label{fig:local_illumination_change}}
\end{figure}
We also reconstruct two objects again under varying light intensity and the results are shown in Fig. \ref{fig:local_illumination_change}, which indicates that local reconstruction is robust against significant brightness variations.
}

\subsection{Object Grasping and Global Reconstruction}
\label{Gripping Experiment}
\begin{figure}[t]
    \centering
     \subfloat[]{
    \includegraphics[width=0.95\linewidth]{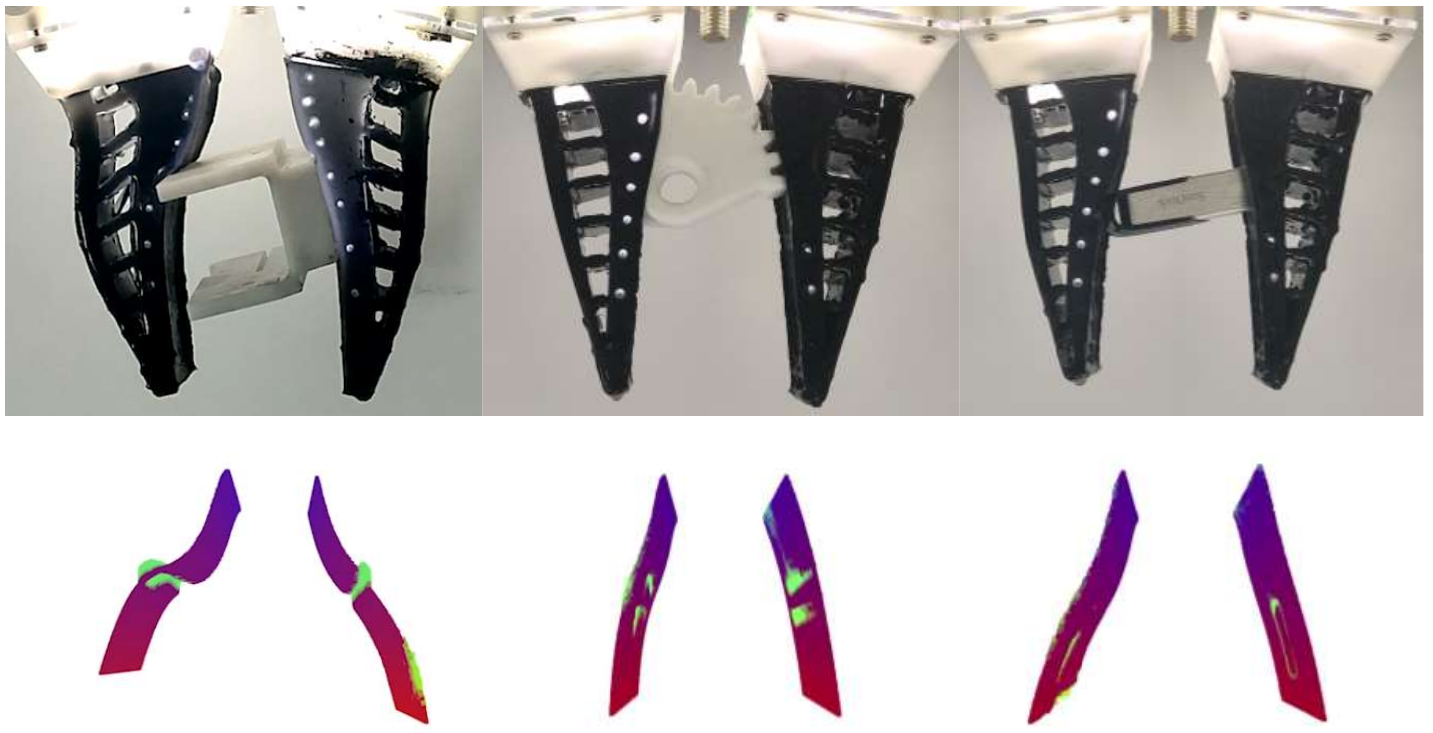}
    \label{fig:grasp}}
    \hfill
     \subfloat[]{
         \includegraphics[width=0.95\linewidth]{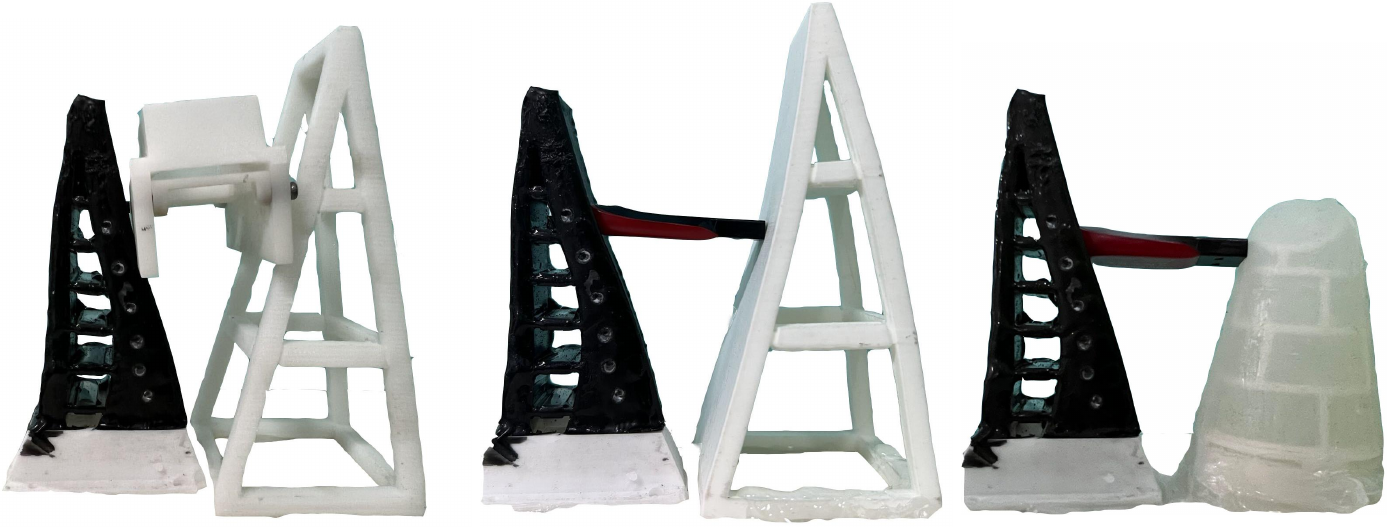}
        \label{fig:grasp_comparison}}
{\caption{(a) Gripper grasping and reconstruction results. The objects include a finger base, a sector gear and a USB stick.(b) Comparison of object grasping performance with existing works\cite{see_thru_finger2}, \cite{guo2024reconstructing}, and \cite{sun2022soft}, presented  in order.}
\label{fig:grips}}
\end{figure}

In this section, we evaluate the gripper's ability to grasp objects of different shapes and sizes and to reconstruct contact faces. 
Fig.~\ref{fig:grasp} illustrates the grasping and reconstruction results.
In order to transform the reconstructed point clouds from two fingers into a unified gripper frame, we model the gripper mechanism and measure the servo length to derive the poses of the two fingers.

Owing to the Fin Ray structure and uni-body elastic material, our finger can deform substantially and compliantly to fit the shape of the objects, resulting in more stable grasping even for challenging shapes such as sector gears. Furthermore, our global and local geometry reconstruction accurately captures the overall size of the grasped object and the detailed contact geometry, such as the two halves of a USB stick socket.

{We further replicate finger designs from \cite{see_thru_finger2,guo2024reconstructing, sun2022soft} and compare object grasping, as shown in Fig.~\ref{fig:grasp_comparison}. Compared to the designs in \cite{see_thru_finger2, guo2024reconstructing}, our finger can localize the contact position and capture fine surface geometry. Compared to the rigid-base sensors like \cite{sun2022soft}, our design offers greater compliance, enabling more stable grasping.}

\subsection{Object Pose Adjustment}
\label{sec:Object Insert}

In this section, we demonstrate the gripper's tactile sensing and pose estimation capabilities for effective object manipulation. For object pose adjustment, the object is received by the gripper in a random orientation. We aim to the gripper to adjust the object's pose to a vertical position based on the tactile feedback received, 
where we first align the camera coordinate frame with the robot arm's end link coordinate frame during mounting and calculate the object's angle to the camera's $z$-axis using PCA. We control the arm to rotate the gripper, along with the grasped object, by the calculated angle to achieve a vertical orientation.

\begin{figure}[]
\centering  
    \centering
    \subfloat[]{
        \includegraphics[width=0.48\linewidth]{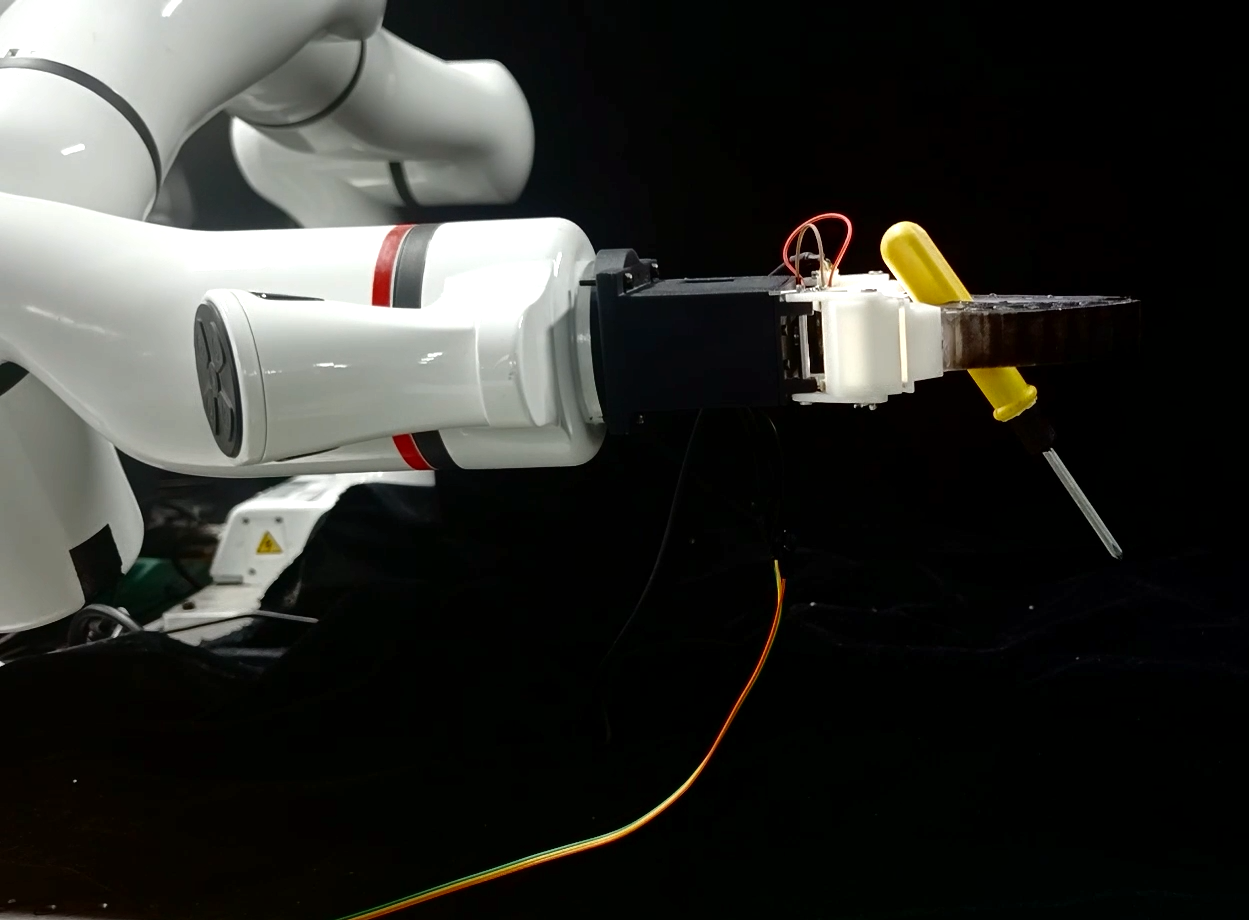}
        \label{fig:root1}
    }
    \subfloat[]{
        \includegraphics[width=0.48\linewidth]{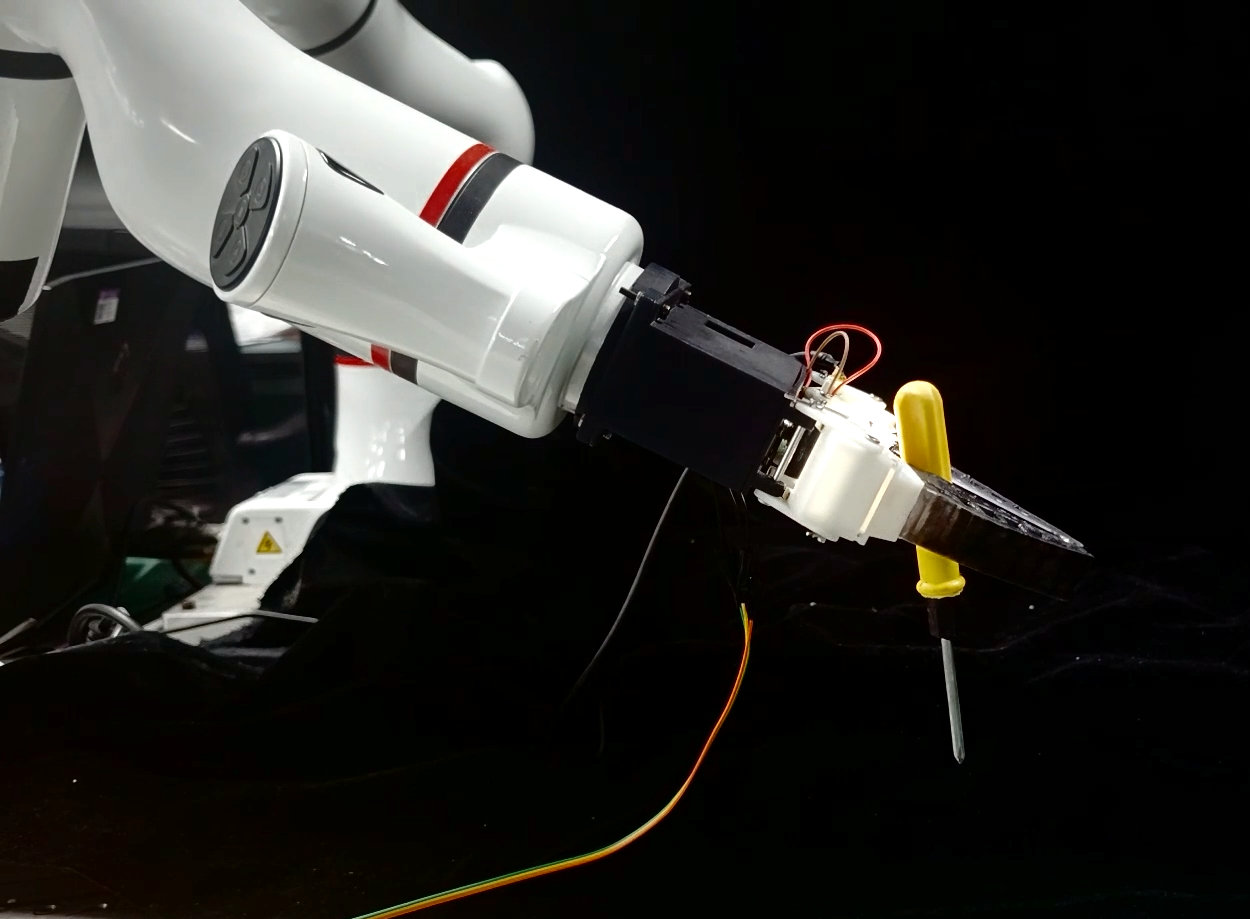}
        \label{fig:root2}
    }

    \centering
    \subfloat[]{
        \includegraphics[width=0.6\linewidth]{
            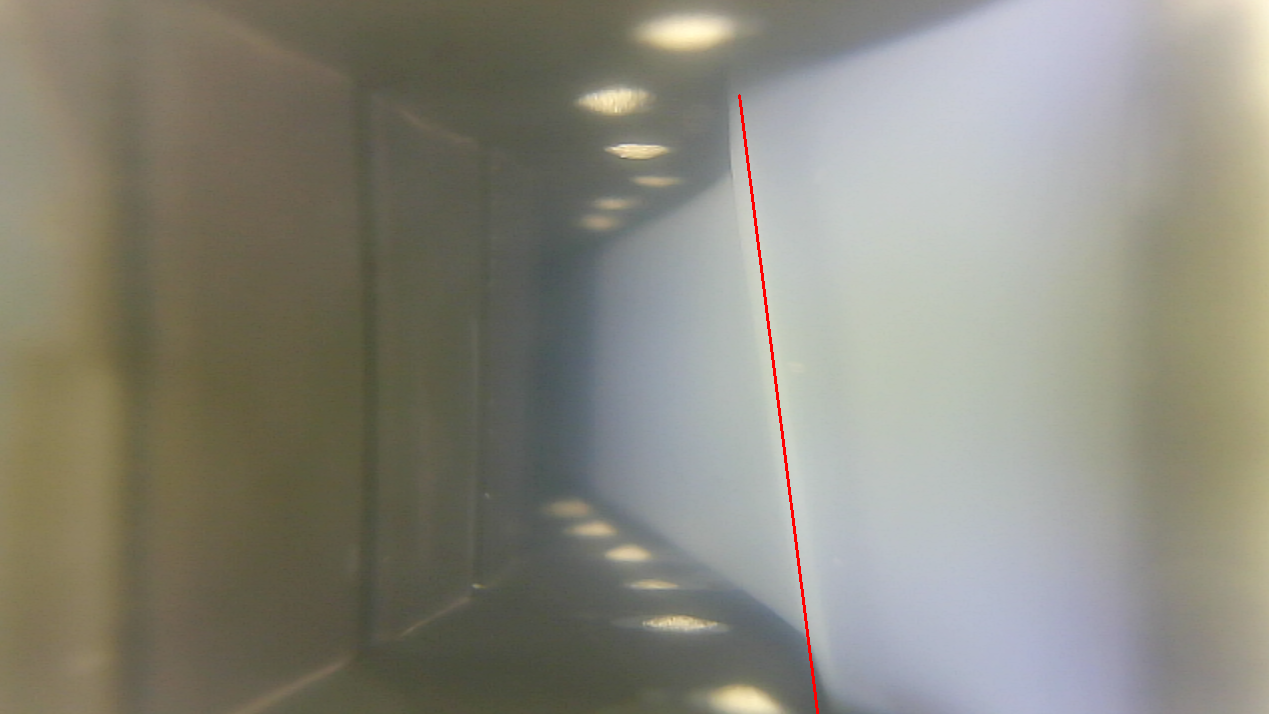
        }
        \label{fig:root3}
    }
\caption{Object pose adjustment experiment. (a) A screwdriver is handed to the gripper at a tilted pose. (b) The arm is controlled to make the screwdriver vertical according to the pose estimation result. {(c) Captured tactile image. The estimated orientation is shown by the red line.}}
\label{fig:tip_experiment}
\end{figure}

Fig.~\ref{fig:tip_experiment} shows the experimental setup and results, where a screwdriver is used in our experiment, and the success criteria is whether the robot is able to insert the tip of the screwdriver into a hole. The insertions succeed for 9 among 10 times when the screwdriver is orientated {toward} the fingertip and 7 among 10 times when orientated {toward} the root. 

\subsection{Computation Time Analysis}

In this section, we evaluate the computation time of each component within our reconstruction pipeline. The experiments are conducted on a desktop computer with an Intel i7-10700 CPU and an NVIDIA GeForce RTX 3090 GPU.

{The image preprocessing step is completed in 2ms.} The combined processes of marker tracking and global reconstruction take approximately 10 ms. Once the tracking results are available, local geometry reconstruction is performed and merged with the global deformation reconstruction to generate the final reconstruction result, which takes 6 ms.

Overall, the entire reconstruction pipeline of a single frame is 18 ms, demonstrating the real-time sensing capability of AllTact Fin Ray.

\section{Conclusion}
\label{sec:conclusion}

In this article, we present AllTact Fin Ray gripper, a compliant, simple-structure gripper embedded visuotactile sensor that can detect contacts from all directions and reconstruct the contact geometries locally and globally. We simplify the fabrication process of the gripper by using uni-body casting and develop a sensing algorithm that only requires white LED illumination and a layer of semi-transparent silicon to accomplish the contact geometry reconstruction. 

{
Our experiments demonstrate that the gripper achieves compliant and stable grasping of objects.The embedded finger sensor can accurately localize the contact positions with an error less than 1 mm, estimate the contact object poses to adjust the pose accordingly, and detect contact forces with a threshold less than 2 N to avoid unexpected contacts during grasping and manipulation. By integrating a lightweight YOLO11 marker detection model, an efficient marker sorting algorithm, and multiprocessing techniques, we reduce the per-frame processing time to 18 ms, enabling real-time sensing capability.
In the future, we plan to investigate materials with enhanced durability. Additionally, we aim to study algorithms for tangential deformation measurement and slip detection, and investigate their applications in advanced manipulation tasks, such as deformable object manipulation and human-robot interaction.
}

\bibliographystyle{IEEEtran}
\bibliography{Mybib.bib}

\vfill

\end{document}